\documentclass[pdflatex,sn-mathphys-num]{sn-jnl}

\usepackage{graphicx}%
\usepackage{multirow}%
\usepackage{amsmath,amssymb,amsfonts}%
\usepackage{amsthm}%
\usepackage{mathrsfs}%
\usepackage[title]{appendix}%
\usepackage{xcolor}%
\usepackage{textcomp}%
\usepackage{manyfoot}%
\usepackage{booktabs}%
\usepackage{algorithm}%
\usepackage{algorithmicx}%
\usepackage{algpseudocode}%
\usepackage{listings}%

\theoremstyle{thmstyleone}%
\theoremstyle{thmstyletwo}%

\theoremstyle{thmstylethree}%

\begin{document}

\title[Article Title]{A Simple State Space Model Excels at Multivariate Time Series Classification}


\author*[1]{\fnm{Hassan} \sur{Saadatmand}}
\email{hassan.saadatmand@monash.edu}

\author[1]{\fnm{Geoffrey I.} \sur{Webb}}

\author[1]{\fnm{Hamid} \sur{Rezatofighi}}

\author[1]{\fnm{Mahsa} \sur{Salehi}}

\affil*[1]{%
  \orgname{Monash University},
  \orgaddress{%
    \city{Melbourne},
    \state{Victoria},
    \country{Australia}
  }
}

\abstract{

Structured state space models (SSMs) have recently emerged as a promising foundation for sequence modeling, with Mamba-based architectures demonstrating strong performance through input-dependent state transitions, albeit at the cost of considerable complexity. However, their application to time-series classification (TSC) has been largely limited to Mamba-style architectures, leaving the broader SSM design space underexplored. We present the first systematic study spanning diagonal SSMs (S4D) and input-dependent SSMs (Mamba family) on large-scale TSC benchmarks, asking whether such architectural complexity is necessary to achieve top performance. Our results reveal a surprising finding: S4D consistently outperforms Mamba-based variants in both accuracy and computational efficiency, challenging the assumption that increased architectural complexity translates to meaningful gains in TSC. Additionally, we introduce MS4—lightweight modifications to S4D via a linear input projection and a simple channel-mixing mechanism—sufficient to improve performance across univariate and multivariate TSC. We further suggest MS4N, a normalized variant that adds a single lightweight normalization layer to stabilize state dynamics and improve training behavior with negligible overhead. We evaluate on 59 datasets across two major benchmarks: MONSTER—with up to 60 million samples, 50K timesteps, and 82 classes—and the commonly used multivariate UEA benchmark, comparing against 15 representative deep and non-deep learning baselines. MS4 and MS4N consistently and substantially outperform Mamba-based models while remaining substantially more efficient and robust, and MS4N achieves matching or superior accuracy to competing deep learning models while those models are much larger (e.g., Mamba- and Transformer-based models that are roughly 2× and 10× larger than MS4N in terms of parameters, respectively). These results underscore that principled, lightweight architectural design is a compelling alternative to scaling complexity, positioning SSMs as a promising foundation for TSC.
}


\maketitle

\section{Introduction}\label{sec1}

Structured State Space Models (SSMs) \cite{gu2022S4} have emerged as a promising family of sequence models for long-range temporal modeling due to their strong theoretical grounding, linear-time complexity, and ability to capture long-range dependencies. Recent progress in sequence modeling has been largely driven by increasingly large and complex models, most notably Transformers \cite{2017Transformer} and input-dependent SSMs such as Mamba \cite{gu2024mamba1}. While these models achieve impressive empirical performance, they do so at the cost of substantial parameter growth, memory usage, and training complexity, which can limit scalability, reproducibility, and practical deployment.

Despite the strong success of structured SSMs, particularly Mamba, in sequence modeling tasks such as natural language processing (NLP) \cite{dao2024mamba2}, their effectiveness in time-series classification (TSC) remains unclear. This is non-trivial, as NLP data are typically discrete and well-structured, whereas time-series data are often continuous, noisy, and non-stationary. These differences raise the question of whether the input-dependent dynamics of Mamba transfer effectively to TSC or introduce unnecessary complexity. To this end, we present the first large-scale evaluation of SSMs for TSC, systematically comparing diagonal SSMs with complex and input-dependent models (Mamba1 \cite{gu2024mamba1}, Mamba2 \cite{dao2024mamba2}) across a diverse set of challenging scenarios, including long sequences, high multivariate inputs, and varying sampling frequencies. Our results provide new insights into the trade-offs between model complexity, efficiency, and accuracy, and help clarify when more complex models are actually needed.

Our results reveal a surprising finding: the simpler diagonal SSM (S4D \cite{gu2022S4}) consistently and significantly outperforms Mamba-based variants in both accuracy and efficiency across a wide range of TSC benchmarks. This challenges the assumption that increased architectural complexity and input-dependent mechanisms always bring benefits across domains. This may be due to S4D’s stronger inductive bias and simpler dynamics, which promote effective modeling of long-range dependencies \cite{Simionato2025}, \cite{gu2020hippo}, and improved generalization, while the added flexibility of input-dependent mechanisms is not always effectively utilized in TSC settings. Motivated by this finding, we revisit S4D and argue that carefully designed diagonal SSMs can achieve strong performance in many TSC settings, without relying on architectural complexity or input-dependent mechanisms \cite{gu2024mamba1}. Our central thesis is that \emph{structured simplicity} provides a scalable and generalizable foundation for TSC, offering not only competitive accuracy but also substantial gains in computational efficiency --- an increasingly important consideration for large-scale and edge-based time-series applications.

This re-examination is timely. Contemporary TSC benchmarks \cite{dempster2025monster, Bagnall2018UEA} increasingly involve 
long temporal horizons (up to tens of thousands of time steps), large sample 
sizes, and deployment scenarios where memory footprint, inference cost, and 
reproducibility are as critical as raw accuracy. While Transformers and 
modern SSMs such as Mamba demonstrate strong performance, their growing 
architectural and computational complexity introduces substantial overhead that 
limits scalability and practical applicability. These trends expose an important 
gap: the lack of strong, simple, and efficient baselines that scale gracefully 
with sequence length and channel count.

Building on S4D, we introduce MS4, which augments diagonal structured 
state space models with two lightweight additions: a learnable input projection 
and a channel-mixing module. The input projection maps raw signals into a 
fixed-dimensional latent space, reducing sensitivity to input dimensionality as 
the number of channels grows. The channel-mixing module introduces controlled 
cross-channel interactions, addressing the limitation of independent channel dynamics in diagonal SSMs while preserving input-independent state transitions and maintaining computational efficiency for TSC. We further propose MS4N, a normalized variant that incorporates a single lightweight normalization layer within the state space block, improving training convergence, accuracy, and robustness while preserving the efficiency and linear-time scaling with respect to sequence length inherent to SSMs.
\begin{figure}[t]
    \centering
    \includegraphics[width=\linewidth]{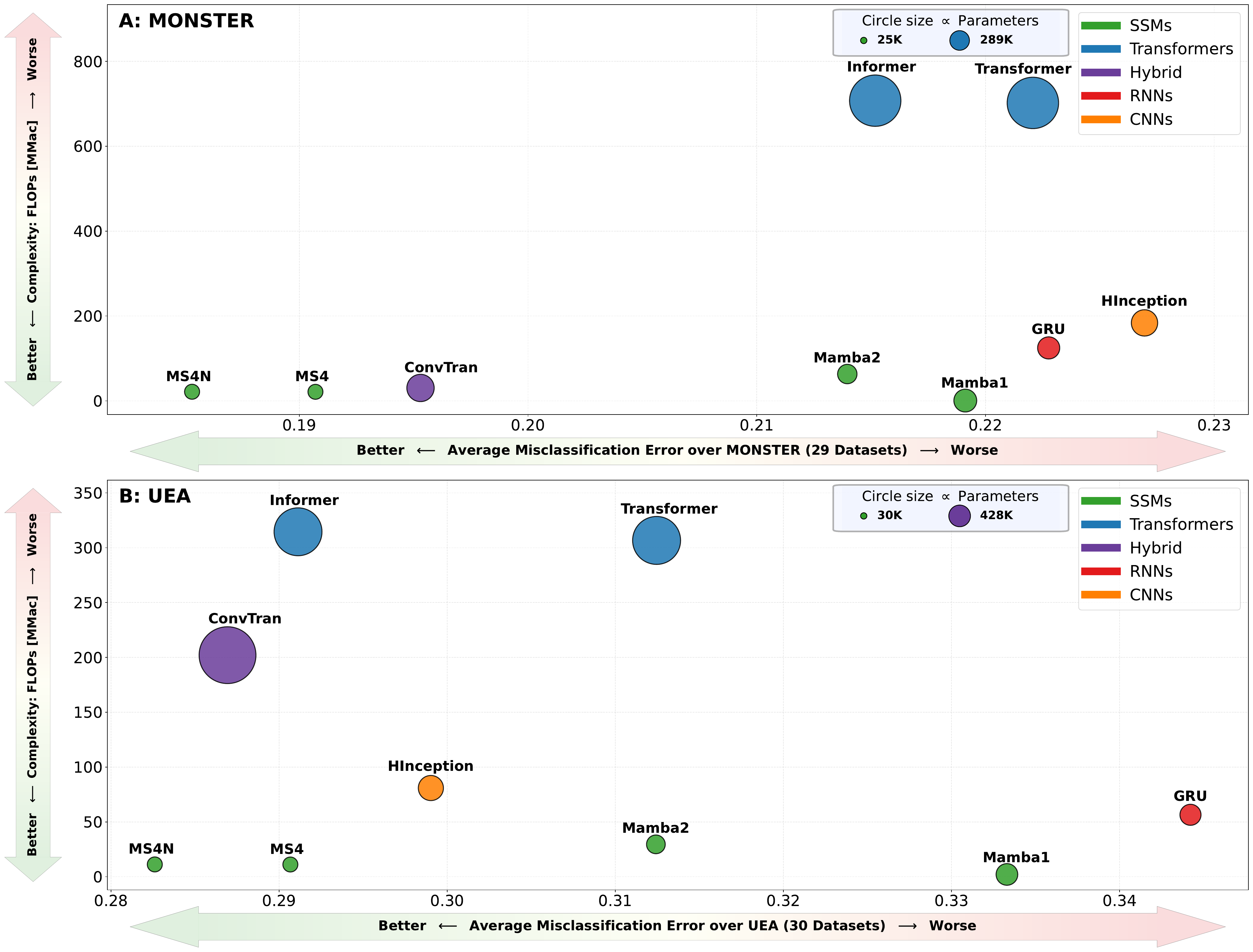}
    \caption{
    Summary plot: accuracy--efficiency trade-off on the MONSTER (A) and UEA (B) 
    benchmarks. Each bubble represents a model, with the x-axis denoting average 
    misclassification error (lower is better), the y-axis denoting computational 
    cost in FLOPs (lower is better), and bubble size proportional to parameter 
    count. Two findings stand out: first, MS4---a simple, underexplored diagonal 
    SSM variant---already surpasses Mamba-based models in both accuracy and 
    computational efficiency, despite its considerably lower complexity. Second, 
    MS4N achieves one of the best overall accuracy--efficiency trade-offs across 
    both benchmarks, outperforming or matching a wide range of models including 
    Mamba-based (Mamba1 \cite{gu2024mamba1}, Mamba2 
    \cite{dao2024mamba2}), Hybrid (ConvTran \cite{Foumani2024ConvTran}), CNNs (HInception \cite{IsmailFawaz2022HinceptionTime}), Transformers (Transformer \cite{2017Transformer}, Informer \cite{zhou2021informer}), and 
     RNNs (GRU \cite{elsayed2019GRUTCS})  architectures, while remaining highly parameter-efficient.
    }
    \label{fig:bubble_overview}
\end{figure}

While the individual elements—projection, channel mixing, and normalization—are standard building blocks, their integration within diagonal SSMs directly addresses three key limitations: sensitivity to input dimensionality, lack of cross-channel interaction, and instability of long-range dynamics. We show that this minimal design is sufficient to achieve strong performance without increasing architectural complexity.

We evaluate our approach on two large-scale TSC benchmarks: the MONSTER repository \cite{dempster2025monster} and the UEA multivariate archive \cite{Bagnall2018UEA}, covering sequence lengths up to 50K, dataset sizes up to 60 million samples, and up to 1,345 channels. These benchmarks span diverse domains and enable rigorous comparison against representative baselines, including Transformers, CNNs, RNNs, Mamba-based architectures, and non-deep learning methods across univariate and multivariate TSC tasks. Figure~\ref{fig:bubble_overview} summarizes the accuracy–efficiency trade-off across all models, from which two key findings emerge. First, MS4—a simple and largely overlooked diagonal SSM—outperforms Mamba-based models in both accuracy and computational efficiency, indicating that increased architectural complexity is not necessary for strong performance in many TSC scenarios. Second, MS4N achieves one of the best overall accuracy–efficiency trade-offs, matching or surpassing a wide range of models while remaining highly parameter-efficient.

Overall, MS4N's advantages are most pronounced in high-dimensional, 
long, high-frequency time-series tasks, while Transformer-based models remain 
competitive on large-scale datasets. The results 
demonstrate that normalized diagonal SSMs provide a simple, efficient, and broadly 
applicable foundation for TSC, achieving strong performance across univariate and 
multivariate settings, and across both deep learning and non-deep learning 
comparisons.

This paper makes the following contributions:

\begin{itemize}

\item We present first systematic study of the SSM family — spanning diagonal SSMs (S4D) and recent input-dependent SSMs — on large-scale TSC benchmarks. We evaluate whether SSMs as a class are competitive in this domain, across diverse settings including long sequences, high channel counts, and varied sampling.

\item We revisit S4D and show that it is a surprisingly strong and underexplored 
baseline for TSC. Contrary to the prevailing trend toward increasingly complex 
architectures, carefully structured and lightweight diagonal SSMs can match or 
surpass substantially larger Transformer- and Mamba-based models.

\item We introduce MS4 and its normalized variant MS4N, extending S4D with a learnable input projection that decouples the model from input dimensionality and controls complexity, a gated channel-mixing mechanism to capture cross-channel dependencies, and a simple normalization layer within the state-space block that significantly improves training stability, convergence speed, and robustness across diverse datasets, while preserving the simplicity and efficiency of S4D and achieving strong performance on multivariate and univariate TSC.

\item Through multiple evaluation criteria—including accuracy, computational complexity, model size (memory efficiency), and performance stability—we conduct extensive experiments on 59 datasets from the MONSTER and UEA benchmarks. MS4N demonstrates a strong overall trade-off, often matching or exceeding substantially larger Transformer- and Mamba-based models while using significantly fewer parameters and lower computational resources. Furthermore, our analysis indicates that the effectiveness of structured SSMs is domain-dependent, with notable advantages in challenging settings such as long sequences, high-frequency signals, and high-dimensional multivariate TSC.

\end{itemize}

\section{Background}
\subsection{Problem definition}

In time-series classification (TSC), each input sequence $x = [x_1, x_2, \ldots, x_L] \in \mathbb{R}^{L \times F}$ is associated with a categorical label $y \in \{1, 2, \ldots, n_c\}$, where $n_c$ is the number of classes and $F$ is the number of features(inputs).  
The objective is to learn a function 
\[
f_{\theta}: \mathbb{R}^{L \times F} \rightarrow \{1, 2, \ldots, n_c\},
\]
that correctly assigns the input sequence to its class based on its temporal dynamics and discriminative patterns.

\subsection{Recent TSC methods}

The landscape of TSC has progressively shifted from classical algorithms toward deep representation learning, although several non–deep learning approaches remain highly competitive. Early benchmarks were largely dominated by ensemble and linear methods; however, the growing demand to process long-range dependency, high-frequency time-series data has driven increased adoption of neural architectures, including RNNs, CNNs, and Transformers. As sequence lengths increase, many existing approaches encounter a fundamental trade-off between effectively modeling long-range temporal dependencies and maintaining computational efficiency. In this work, we review the most representative models across these paradigms, highlighting their strengths, limitations, and scalability characteristics.

\subsubsection{Ensemble and convolutional feature-based approaches}
Historically, ensemble frameworks such as COTE \cite{Bagnall2015COTE} established strong baselines for TSC by combining diverse representations, including interval-based, spectral, and shapelet-based features \cite{He2019ShapeletEnsemble}. Their strong performance arises from aggregating many heterogeneous classifiers, each of which is itself computationally expensive. Consequently, these methods scale poorly with increasing sequence length and dataset size, limiting their applicability in large-scale or real-time scenarios.

To address these limitations, a class of more scalable methods based on convolutional feature transforms has emerged. These methods belong to a broader family of approaches that first map input time series into very high-dimensional feature spaces and then learn a linear mapping from these representations to the target classes. Within this paradigm, ROCKET \cite{Dempster2020ROCKET} and its variants, MiniROCKET \cite{Dempster2021MiniRocket}, MultiROCKET \cite{Tan2022MultiROCKET} and Hydra \cite{Dempster2023Hydra} use large collections of random convolutional kernels as feature extractors \cite{saad2025multiobjective,azadi2026multiobjective}, followed by classification with a simple linear model. Related dictionary-, interval- and transform-based classifiers include WEASEL~2.0 \cite{schafer2023weasel2}.

More recently, Hydra \cite{Dempster2023Hydra} and Quant \cite{Dempster2024Quant} extend this convolutional feature-based paradigm by incorporating more expressive aggregation mechanisms, such as grouped kernel counting and tree-based models, while retaining the efficiency benefits of randomized or multi-scale convolutional transforms. Although these methods improve scalability relative to classical ensemble frameworks, they still rely on aggregating large numbers of transformed features or weak learners, which can limit performance on very long sequences.
and Quant \cite{Dempster2024Quant}.

\subsubsection{Recurrent and convolutional neural networks}
Deep learning introduced RNNs and their gated variants (LSTMs \cite{YU2021RNNs}, GRUs \cite{elsayed2019GRUTCS} ) to explicitly model time series problems. However, their inherently sequential nature prevents parallelization and often leads to vanishing gradients over long horizons. Conversely, CNNs, such as Fully Convolutional Networks (FCN) \cite{2017FCN} and InceptionTime \cite{Fawaz2020InceptionTime} are convolutional-based neural network for TSC, inspired by the Inception architecture.  Multiple layers of convolutional kernels are applied to transform the input time series and relies on global pooling, without explicitly modeling temporal location information. H-InceptionTime \cite{IsmailFawaz2022HinceptionTime} extends this approach by combining learned and fixed convolutional kernels to enhance feature extraction. Disjoint-CNN \cite{Disjoint-CNN} shows that decomposing 1D convolution kernels into disjoint temporal and spatial components improves accuracy with minimal computational overhead. Yet, CNNs are constrained by a finite receptive field, requiring significantly deeper architectures or dilated kernels to capture the long-range dependencies inherent in complex time-series signals.

\subsubsection{Attention-based and hybrid models}
The introduction of self-attention mechanisms enabled direct modeling of long-range dependencies by computing pairwise interactions across all timesteps. The vanilla Transformer \cite{2017Transformer} with appropriate positional encodings can effectively capture temporal dependencies in multivariate time series. However, standard Transformers lack explicit temporal inductive biases, leading to sensitivity toward high-frequency noise common in sensor data.
To address multivariate time-series structure, specialized attention mechanisms have been proposed. Cross Attention Stabilized Fully Convolutional Neural Network (CA-SFCN) \cite{ijcai2020CA-SFCN} combines FCN with dual self-attention modules—temporal attention for long-range dependencies and variable attention for cross-channel interactions. Gated Transformer networks \cite{liu2021gated} employ a two-tower architecture where separate attention streams process temporal and channel information, merged through learnable gating.

Beyond classification-specific architectures, several general-purpose time-series Transformers originally designed for forecasting have demonstrated strong transferability to classification tasks. Informer \cite{zhou2021informer} introduced ProbSparse self-attention to reduce the quadratic complexity of standard attention, enabling efficient processing of long sequences. PatchTST \cite{PatchTST} applies a patching strategy that segments time series into subseries-level patches, treating each patch as a semantic token analogous to words in NLP, thereby reducing the effective sequence length while preserving temporal information. iTransformer \cite{liu2023itransformer} inverts the conventional Transformer architecture by applying attention across variates rather than time steps, effectively treating each time series channel as a token to capture multivariate correlations.

Despite these innovations, Transformers face fundamental limitations. Standard self-attention incurs quadratic complexity $O(L^2)$ relative to sequence length and scales poorly with high channel dimensionality \cite{gu2024mamba1,ZHOU2023TransProblem}. Furthermore, they lack inherent sequence awareness, necessitating external positional encodings \cite{Foumani2024ConvTran}. While efficient variants like Informer \cite{zhou2021informer} or FlashAttention \cite{dao2022flashattention} mitigate these costs, they still impose high memory demands and often require massive datasets to generalize \cite{ma2024survey_pre}. For long-sequence problems, these overheads render Transformers impractical, motivating the shift toward linear-time structured state-space models.

Hybrid architectures represent a promising solution by attempting to leverage the complementary advantages of both convolutional local feature extraction and Transformer-based global modeling. A prominent example is ConvTran \cite{Foumani2024ConvTran}, a state-of-the-art model that demonstrates that learned convolutional features are more effective than raw inputs for self-attention. However, this dual-stage design inherits the quadratic complexity of Transformers alongside the significant parameter overhead of deep CNNs. Consequently, the approach suffers from heavy memory demands that scale poorly with high dimensionality, creating a computational bottleneck for the high-dimensional benchmarks.

\subsection{Evolution of structured SSMs}

Structured SSMs \cite{gu2022S4} originate from classical control theory, representing dynamical systems through coupled first-order differential equations. Unlike fully "black-box" sequence models, SSMs impose a mathematical structure that facilitates long-range dependency modeling with linear-time complexity. The continuous-form of a Linear Time Invariant (LTI) system is defined by:
\begin{align}\dot{h}(t) &= A h(t) + B x(t),\\
y(t) &= C h(t) + D x(t),\end{align}
where $A, B, C, D$ are system matrices governing internal dynamics, input influence, output mapping, and skip connection, respectively. To process digital time-series $x \in \mathbb{R}^{L \times F}$, $F$ is the number of features/channels and $L$ is the input sequence length, these equations are discretized (e.g., via Bilinear or Zero-Order Hold (ZOH) transforms) into a recurrent form: \begin{align} h_k &= \bar{A} h_{k-1} + \bar{B} x_k, \\ 
y_k &= C h_k + D x_k, \end{align}

\subsubsection{The diagonal SSM}
The Structured State Space Sequence model (S4) \cite{gu2022S4} addressed the vanishing/exploding gradient problem in SSMs by initializing $A$ with the HiPPO framework \cite{gu2020hippo}, which projects history onto orthogonal polynomials. While S4 utilized a complex ``Normal Plus Low-Rank'' parameterization to enable $O(L \log L)$ training via the Fast Fourier Transform (FFT), subsequent research sought to simplify this structure. Notably, S4D (Diagonal S4) \cite{gupta2022diagonal} demonstrated that constraining $A$ to a purely diagonal matrix matches the performance of more complex parameterizations while significantly reducing implementation difficulty. 

In the standard LTI formulation, the state matrix $A \in \mathbb{R}^{N \times N}$ ($N$ is the state space dimension) and the input matrix $B \in \mathbb{R}^{N \times 1}$ (for a single channel and one step time) govern the latent dynamics. The core innovation of S4D is the diagonalization of $A$, which decouples the hidden state $h(t) \in \mathbb{R}^{N}$ into $N$ independent scalar systems. Rather than utilizing a dense matrix $A \in \mathbb{R}^{N \times N}$, the state evolution is defined through a vector of complex eigenvalues $\Lambda = (\lambda_1, \lambda_2, \dots, \lambda_{N/2}) \in \mathbb{C}^{N/2}$. To satisfy the requirement for real-valued outputs, complex-conjugate symmetry is employed, allowing $N/2$ complex parameters to represent an $N$-dimensional dynamical system. Each eigenvalue is parameterized as:
\begin{equation}
\lambda_n = -\exp(a_n) + i \cdot b_n, \quad a_n, b_n \in \mathbb{R}
\end{equation}
The constraint $\text{Re}(\lambda_n) < 0$ ensures that the system is stable, providing a fading memory that remains numerically stable over long temporal horizons.

\subsubsection{Discretization and convolution} 
To process discrete time-series data, the continuous system $(\Lambda, B, C)$ must be mapped to its discrete-time counterpart $(\bar{A}, \bar{B}, C)$ relative to a learnable step size $\Delta \in \mathbb{R}^+$. Applying Zero-Order Hold (ZOH) \cite{Pechlivanidou2022ZOH} transformation to the diagonalized system results in element-wise discrete parameters:
\begin{align}
\bar{A} &= \exp(\Delta \Lambda) \\
\bar{B} &= \Lambda^{-1}(\bar{A} - I)B
\end{align}
where $I$ denotes the identity. The initial values for $\Lambda$ are not random; they are derived from the HiPPO-LegS \cite{gu2020hippo} matrix. By initializing $\Lambda$ such that it approximates the eigenvalues of the HiPPO matrix, the model starts with a mathematically optimal ``memory'' that approximates the input history using orthogonal Legendre polynomials. This ensures that even before training begins, the diagonal SSM is biased toward preserving historical information.

The discrete recurrence $h_k = \bar{A} h_{k-1} + \bar{B} x_k$ and $y_k = C h_k$ (only consider the state output term) can be represented as a global convolution. The SSM kernel $\mathcal{K} \in \mathbb{C}^L$ for one channel is computed as: 
\begin{align}
\quad \mathcal{K} = \big( C \bar{A}^0 \bar{B},\ C \bar{A}^1 \bar{B},\ \dots,\ C \bar{A}^{L-1} \bar{B} \big)
\end{align}
In a general SSM with a dense matrix, computing the sequence of powers $(\bar{A}^0, \bar{A}^1, \dots, \bar{A}^{L-1})$ presents a significant computational bottleneck for calculating the kernel $\mathcal{K}$. By employing a diagonal structure, this operation simplifies to element-wise scalar exponentiation. The state output $y_s$ is then obtained by the convolution of the input $x$ and the kernel $\mathcal{K}$ via FFT, which reduces the computational complexity from quadratic to nearly linear:\begin{equation}y_s = \mathcal{K} * x = \mathcal{F}^{-1} \big( \mathcal{F}(\mathcal{K}) \odot \mathcal{F}(x) \big)\end{equation}where $\mathcal{F}$ denotes the FFT and $\odot$ denotes element-wise multiplication, and $\mathcal{F}^{-1}$ is inverse FFT. The transition from the recurrent form to the convolutional representation is not merely a mathematical convenience; it allows Structured SSMs to bypass the sequential bottleneck that plagues traditional RNNs. The detailed procedure for this parallel forward pass via the S4D kernel is outlined in Algorithm 1.
\begin{algorithm}[tbp]
\caption{S4D Algorithm}
\begin{algorithmic}[1]
\Require Input sequence $x \in \mathbb{R}^{L \times F}$;
         learnable parameters $(\Lambda, B, C, D, \Delta)$
\Ensure output $y \in \mathbb{R}^{L \times F}$
\vspace{0.3em}

\State $\mathcal{K} = \text{S4D\_Kernel}(\Lambda, B, C, \Delta)$
      \Comment{Kernel $\mathcal{K} \in \mathbb{R}^{L \times F}$, Eq. 6-8}

\State $y_s = \mathcal{F}^{-1} \big( \mathcal{F}(x) \odot \mathcal{F}(\mathcal{K}))$
      \Comment{Diagonal SSM convolution via FFT}

\State $y = y_s + x \odot D$
      \Comment{Feedthrough / skip term}

\State $y = \mathrm{Dropout}(\mathrm{GELU}(y))$
      \Comment{Nonlinearity + regularization}

\State \Return $y$
\end{algorithmic}
\end{algorithm}
During training, the entire input sequence of length $L$ is available. By leveraging the convolutional formulation, the model computes state updates for all timesteps in parallel. The computational bottleneck is the FFT and its inverse, resulting in a complexity of $O(L \log L)$ per channel. In contrast, Transformer models based on self-attention incur $O(L^2)$ complexity with respect to sequence length.

While the convolutional form is efficient for training, SSMs can be seamlessly reverted to its recurrent form for inference. At test time, the model does not need to re-process the entire history to predict the next state; it only requires the current input $x_k$ and the previous hidden state $h_{k-1}$. This per-step computation is independent of the sequence length $L$, resulting in constant-time ($O(1)$) inference and constant-memory ($O(1)$) overhead. This makes SSMs uniquely suited for ``online'' or streaming applications—such as real-time medical monitoring or industrial sensor analysis—where low-latency processing of long-duration signals is mandatory.

\subsubsection{The shift toward input-dependent SSMs}

Recent progress in SSMs modeling has increasingly favored input-dependent dynamics. Mamba \cite{gu2024mamba1} represents a major departure from classical linear time-invariant SSMs by introducing a selective mechanism in which the state transition matrix, input projection, and discretization step size are conditioned on the input at each timestep. This design enables the model to dynamically modulate its temporal behavior, effectively introducing an attention-like mechanism while preserving linear-time complexity. Mamba-2 \cite{dao2024mamba2} further advances this paradigm through State Space Duality (SSD), which unifies recurrent state updates with parallelizable computation to improve hardware efficiency.

Motivated by the strong performance of Mamba in discrete domains such as natural language processing, several recent works have explored its applicability to TSC. TSCMamba \cite{AHAMED2025TSCmamba} integrates Mamba blocks within a multi-view fusion framework to capture complementary temporal representations, while Jiang et al.~\cite{Jiang2025Mamba-hypergraph} incorporate Mamba-based dynamics into hypergraph neural networks to model higher-order relationships among time-series channels. These architectures have shown promising results in medical biosignal analysis, including EEG-based dementia detection \cite{le2024SSMeeg} and ECG \cite{mehari2022ecgssm}, where modeling long-range physiological dependencies is critical.

Despite these successes, the shift toward input-dependent SSMs introduces notable challenges when applied to continuous-valued time series. Unlike discrete token sequences, time-series data are often characterized by high-frequency noise and stochastic fluctuations, which can cause selective mechanisms to respond to transient variations rather than stable temporal patterns. To mitigate this, recent Mamba-based models frequently rely on additional architectural components—such as selective gating, wavelet transforms, or mixture-of-experts—to stabilize training, increasing both computational complexity and memory overhead. As a result, the benefits of input-dependent dynamics may diminish in large-scale or long-horizon TSC settings.

In this work, we argue that the inductive bias of input-invariant diagonal SSM remains highly effective for multivariate TSC. By revisiting the diagonal SSM (S4D) and augmenting it with lightweight normalization and explicit channel-mixing, we demonstrate that strong accuracy and scalability can be achieved without the overhead of data-dependent state transitions. This challenges the prevailing assumption that increased complexity is necessary for effective long-range modeling.

\section{Methodology}

Building on the theoretical foundation of diagonal structured SSMs, we introduce MS4, an extension of S4D augmented with two lightweight components: a learnable input projection and a channel-mixing module. We further propose MS4N, a normalized variant that incorporates a single LayerNorm operation to improve accuracy, training stability, and robustness. 

These modifications are motivated by key challenges in multivariate TSC, including cross-channel dependencies, high-frequency noise, and numerical instability over long temporal horizons. Rather than increasing architectural complexity, our goal is to introduce the minimal set of inductive biases required to address these limitations while preserving the efficiency and scalability of diagonal SSMs.

\begin{figure}[htbp]
    \centering
    \includegraphics[width=0.95\textwidth]{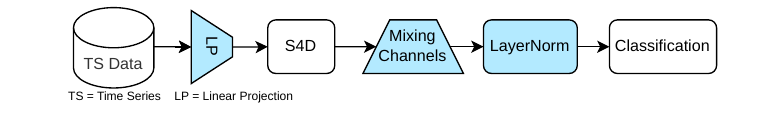}
    \caption{The proposed MS4N architecture for multivariate TSC.}
    \label{fig:MS4M}
\end{figure}

The overall architecture is illustrated in Figure~\ref{fig:MS4M}, and the forward pass for both MS4 and MS4N is summarized in Algorithm~\ref{alg:ms4n}.

\subsection{The Multivariate extension (MS4)}

Standard diagonal SSMs such as S4D model each latent dimension independently and are not inherently designed to capture inter-channel correlations, which are critical in multivariate domains such as medical biosignals and sensor networks. MS4 addresses this limitation by introducing two lightweight modules alongside the S4D block.

\begin{algorithm}[tbp]
\caption{MS4 and MS4N forward pass}
\label{alg:ms4n}
\begin{algorithmic}[1]
\Require Input sequence $x \in \mathbb{R}^{L \times F}$; parameters $\{W_i, b_i\}_{i=1}^{4}$, $(\Lambda, B, C, D, \Delta)$
\Ensure Class logits $z \in \mathbb{R}^{n_c}$

\Statex \textit{// Input projection}
\State $x_p \gets x W_1 + b_1$ \Comment{$F \rightarrow H$}

\Statex \textit{// Core S4D block}
\State $y \gets \text{S4D}(x_p, \Lambda, B, C, D, \Delta)$

\Statex \textit{// Gated channel mixing}
\State $y_2 \gets y W_2 + b_2$ \Comment{$H \rightarrow 2H$}
\State Split $y_2$ into $[a, b]$
\State $g \gets a \odot \sigma(b)$

\Statex \textit{// Normalization (MS4N only)}
\If{MS4N}
    \State $g \gets \textsf{LayerNorm}(g)$
\EndIf

\Statex \textit{// Classification head}
\State $h \gets \text{GlobalAveragePooling}(g)$
\State $z \gets \text{GELU}(h W_3 + b_3)W_4 + b_4$

\State \Return $z$
\end{algorithmic}
\end{algorithm}

\subsubsection{Learnable input projection}

Given an input sequence $x \in \mathbb{R}^{L \times F}$, we first apply a learnable linear projection:
\begin{equation}
x_p = x W_1 + b_1
\end{equation}
where $W_1 \in \mathbb{R}^{F \times H}$. 

This projection serves as a lightweight feature extractor that maps raw inputs into a latent space of dimension $H$. By decoupling the model from the original channel dimensionality $F$, it enables scalability to high-dimensional inputs while attenuating high-frequency noise before temporal modeling.

\subsubsection{Temporal representation learning}

The projected sequence $x_p$ is processed by the S4D block, which performs temporal modeling using diagonal state-space kernels. To preserve signal fidelity and improve gradient flow, a direct feedthrough connection is applied:
\begin{equation}
y = y_s + D \odot x_p
\end{equation}
where $y_s = \mathcal{K} * x_p$ denotes the convolutional state output.

This stage encodes global temporal dependencies, where each latent dimension captures a distinct dynamical mode. The structured, time-invariant nature of S4D provides a stable global receptive field and robustness to transient noise.

\subsubsection{Gated channel mixing}

To capture cross-channel dependencies, we introduce a gated channel-mixing module:
\begin{align}
y_2 &= y W_2 + b_2 \\
[a, b] &= \text{Split}(y_2) \\
g &= a \odot \sigma(b)
\end{align}

This Gated Linear Unit (GLU) enables selective interaction between latent dimensions, providing expressive cross-channel modeling without incurring the quadratic complexity of attention mechanisms.

\subsection{State normalization (MS4N)}

A key challenge in long-sequence SSMs is the accumulation of numerical instability in state dynamics. To address this, MS4N introduces a lightweight normalization step:
\begin{equation}
g = \textsf{LayerNorm}(g)
\end{equation}

This operation acts as a dynamic range stabilizer, controlling the scale of hidden representations and improving gradient flow. As a result, MS4N achieves faster convergence and improved robustness, while preserving the $O(L \log L)$ training and $O(1)$ inference properties of diagonal SSMs (see Section 5.1.7).

\subsection{Design rationale: minimal modifications to S4D}

Although the components used in MS4N are individually standard, their role within SSMs is fundamentally different from their use in conventional architectures. Diagonal SSMs exhibit three key limitations: (1) lack of cross-channel interaction, (2) sensitivity to input dimensionality, and (3) instability in long-range state evolution.

Each component in MS4N addresses one of these limitations. The input projection stabilizes and compresses high-dimensional inputs, the channel-mixing module introduces controlled inter-channel interactions, and the normalization layer ensures stable state dynamics over long sequences. Together, these modifications form a minimal and principled design that overcomes the structural limitations of diagonal SSMs without significantly increasing computational complexity.

\subsection{Understanding MS4N's effectiveness}

The effectiveness of MS4N arises from the alignment between its structure and the statistical properties of real-world time-series data. Diagonal SSMs can be interpreted as a bank of learnable linear filters, where each latent dimension captures global temporal dependencies with a specific frequency response. Since many real-world signals (e.g., audio, EEG, sensor data) are dominated by smooth, low-frequency components, these filters provide a strong inductive bias by capturing long-range dependencies while suppressing high-frequency noise.

In contrast, input-dependent models such as Mamba dynamically adapt their parameters at each timestep. While this increases expressivity, it can also lead to sensitivity to local fluctuations and noise. By enforcing input-invariant dynamics, MS4N implicitly regularizes temporal structure, improving robustness and generalization. The addition of normalization further stabilizes training by controlling representation scale and improving gradient propagation. Together, these properties demonstrate that carefully structured and lightweight models, when aligned with data characteristics, can match or surpass more complex architectures.

\subsection{Complexity analysis}

The input projection maps $F$-dimensional inputs to a fixed hidden dimension $H$, 
contributing $\mathcal{O}(LFH)$ operations and decoupling all subsequent computation 
from $F$. The FFT-based SSM kernel adds $\mathcal{O}(L \log L)$, and the gated 
channel mixing contributes $\mathcal{O}(LH^2)$, giving a total per-block complexity 
of $\mathcal{O}(LFH + LH^2 + L \log L)$. The FFT term is negligible in practice, so either $\mathcal{O}(LFH)$ or 
$\mathcal{O}(LH^2)$ dominates depending on whether $F \gg H$ or $H \gg F$. In 
all regimes, complexity remains linear in $L$, contrasting with the $\mathcal{O}(L^2)$ 
scaling of standard self-attention deep learning models for TSC. Fixing $H$ independently of $F$ further ensures 
that models trained on one sensor configuration generalise to others without 
any change in downstream computational cost.

During inference, MS4N updates its hidden state as $h_k = f(h_{k-1}, x_k)$, 
retaining only $h_{k-1} \in \mathbb{R}^H$ and $x_k$ at each step. Combined with 
an $\mathcal{O}(H^2)$ classification head that is independent of $L$, the total 
inference memory footprint is $\mathcal{O}(H)$ — constant in both $L$ and $F$, 
and in stark contrast to the $\mathcal{O}(L)$ KV-cache of autoregressive 
Transformers.
\section{Experimental setup}

\subsection{Datasets}
For our comprehensive experimental evaluation, we consider two major and widely used benchmarks in TSC, with particular emphasis on MONSTER repository~\cite{dempster2025monster}, a large-scale benchmark specifically designed to evaluate the scalability and robustness of deep learning–based models on massive datasets and long sequences. MONSTER comprises 29 datasets spanning a diverse range of domains including audio, sensor, motion, and medical signals, with sequence lengths and sample sizes that are orders of magnitude larger than those found in conventional benchmarks, making it particularly challenging for models with high computational overhead.

In addition, we evaluate on the UEA multivariate time-series archive~\cite{Bagnall2018UEA}, which contains 30 multivariate datasets covering a broad spectrum of real-world classification tasks such as human activity recognition, electroencephalography, and motion capture, providing a rigorous testbed for assessing a model's ability to capture inter-channel dependencies across time. Collectively, these two benchmarks span a total of 59 datasets across a wide range of application domains, sequence lengths, sample sizes, and levels of multivariate complexity, providing a rigorous and diverse testbed for assessing model performance, scalability, and generalization across TSC tasks.

\subsection{Baselines}
Overall, we compare against 15 baseline methods selected to provide comprehensive coverage of the time-series modeling landscape, spanning recent state-of-the-art approaches, established representative models, and both deep learning (DL) and non–deep learning (non-DL) paradigms. 

Our deep learning baselines include ConvTran \cite{Foumani2024ConvTran}, a recent hybrid architecture combining convolutional neural networks and Transformers; convolutional models such as HInception \cite{IsmailFawaz2022HinceptionTime} and FCN \cite{2017FCN}; Transformer-based models including the vanilla Transformer \cite{2017Transformer}, Informer \cite{zhou2021informer}, and iTransformer \cite{liu2023itransformer}; and recurrent architectures, namely LSTM \cite{hua2019lstmTS} and GRU \cite{chung2014GRU}. Together, these models represent the most widely adopted deep learning paradigms for time-series modeling. To evaluate structured sequence modeling approaches, we further include state-space models (SSMs), namely S4/MS4 \cite{gu2022S4}, Mamba-1 \cite{gu2024mamba1}, and Mamba-2 \cite{dao2024mamba2}, which capture recent advances in efficient long-sequence learning. In addition, we consider three representative non-DL methods: Quant \cite{Dempster2024Quant}, Hydra \cite{Dempster2023Hydra}, and Extremely Randomised Trees (ET) \cite{geurts2006ET}. These ensemble and feature-based approaches are included due to their strong empirical performance and complementary inductive biases. Collectively, this diverse set of baselines enables a fair and thorough comparison across architectural families, model complexities, and learning paradigms.

\begin{figure}[htbp]
    \centering
    \includegraphics[width=1\textwidth]{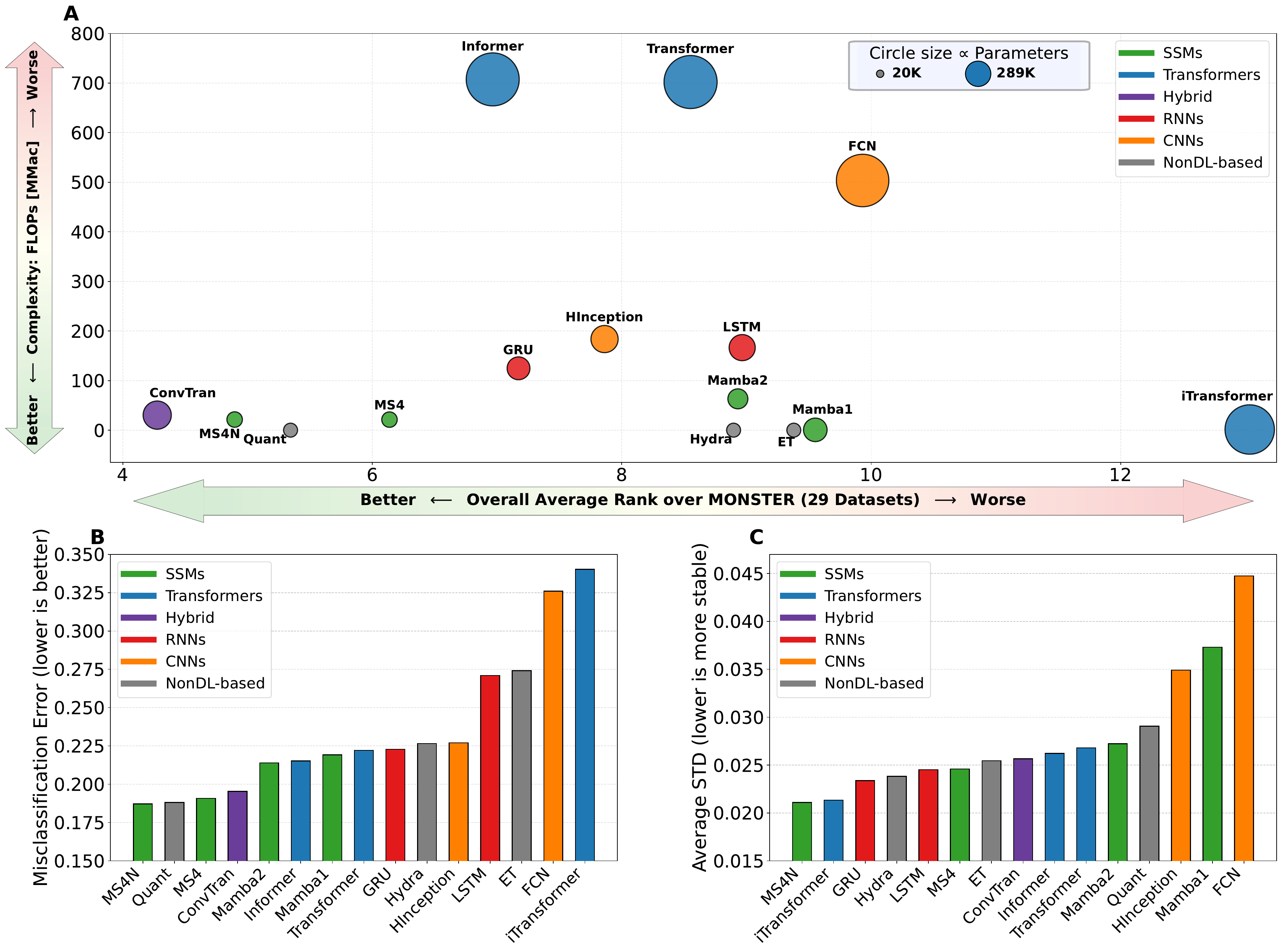}
    \caption{Three-panel comparison of all models over 29 MONSTER datasets. (A) Bubble plot of overall average rank versus computational cost (FLOPs in MMac); bubble size is proportional to the number of parameters---models in the bottom-left
    are both accurate and efficient. NonDL-based models (Quant, Hydra, ET) assigned 20K parameters and 1 FLOP as proxy values. (B) Average misclassification error per model,
    sorted in ascending order. (C) Average standard deviation (STD) per
    model, sorted in ascending order---lower values indicate more stable training.
    All panels are colored by architectural family: SSMs (green), Transformers (blue), Hybrid (purple), RNNs (red), CNNs (orange), and NonDL-based (gray)}
    \label{fig:MonsterOverallFlops}
\end{figure}

\subsection{Implementation details}
All deep learning models were trained using the Adam optimizer with an initial learning rate of 0.001. Unless otherwise stated, we adopted a unified training protocol across all architectures to ensure a fair comparison. For MONSTER, the batch size is set to 256. The maximum number of training epochs is set to 200. The AudioMNIST dataset is the only exception, where a reduced batch size of 64 is used due to GPU memory constraints. For UEA datasets, the maximum number of epochs is set to 300, with batch sizes ranging from 4 to 64 depending on dataset size.

To promote generalization and prevent overfitting, early stopping was employed based on a validation set constructed from 10\% of the training data. Model selection was performed by choosing the checkpoint achieving the best validation performance. Using the original five-fold cross-validation splits, we report the average performance across all folds.

\subsection{Evaluation Metrics}

We evaluate all methods along four complementary dimensions: misclassification error, defined as the fraction of incorrectly classified instances ($\text{Error} = 1 - \text{Accuracy}$); average rank, computed by ranking all models per dataset by ascending error (rank~1 = best) and reporting the mean across datasets, which provides a distribution-free, outlier-robust summary of relative standing; computational complexity, reported as the number of multiply--accumulate operations (FLOPs, in MMac) for a single forward pass and the total number of trainable parameters, both obtained via the \textit{ptflops} library on a fixed representative input, with non-deep learning baselines (Quant, Hydra, ET) assigned nominal proxy values (1 FLOP, 20K parameters) for visualization; and training stability, measured as the standard deviation (STD) of misclassification error across folds or runs, where lower STD indicates more consistent training behavior across diverse datasets.

\section{Results and discussion}
This section presents the experimental results and provides a comprehensive analysis of the proposed approach. We begin by evaluating the overall performance–complexity trade-off of MONSTER and UEA, followed by a detailed discussion of results across individual domains and datasets.

\subsection{Overall performance--complexity trade-off on MONSTER}

Figure~\ref{fig:MonsterOverallFlops} compares models across five architectural
families: SSMs (green), Transformers (blue), RNNs (red), CNNs (orange), and
non-deep-learning (NonDL) methods (grey) through three complementary views:
Panel~A shows predictive rank against computational cost (MMac) and model size;
Panel~B reports average misclassification error; and Panel~C reports average
standard deviation (STD), capturing cross-validation stability. For visualization,
NonDL models are assigned 1 FLOP and 20K parameters—below the minimum among
deep models—serving as a conservative efficiency baseline.

A consistent pattern emerges across all three panels: explicit temporal inductive
bias appears to be more important than increasing model scale in continuous TSC.
Models that impose structure aligned with the data characteristics (e.g. MS4N and MS4) achieve strong
accuracy, efficiency, and stability, while more expressive architectures (e.g. FCN, Informer, and Transformer) often
incur higher computational cost without proportional gains. These observations
suggest that structured simplicity, rather than architectural complexity, appears to be an important factor in achieving robust performance in this setting.

Within the SSM family, MS4 outperforms more complex selective SSM variants. Despite its simplicity,
MS4 (rank 4th, error 0.191) achieves better accuracy, lower computational cost,
and improved stability compared to both Mamba1 and Mamba2. Notably, Mamba-2
requires approximately $3\times$ the FLOPs and nearly $2\times$ the parameters
of MS4, yet yields lower accuracy and higher variance. This may indicate that
input-dependent state transitions introduce additional flexibility that is not
always beneficial for the relatively smooth and locally correlated signals in
these datasets, whereas fixed structured dynamics provide a more appropriate
inductive bias.

MS4N achieves the best accuracy  or lowest
error (0.187) among all 15 methods. Importantly, MS4N also attains the lowest
standard deviation ($\approx$0.021), indicating highly stable training dynamics.
These improvements are achieved with only 25K parameters and 21.5~MMac—the
smallest footprint among all deep learning models—demonstrating that a simple
normalization mechanism can significantly enhance both stability and predictive
performance without increasing model complexity.

Among deep learning competitors, ConvTran emerges as the most effective hybrid
architecture, combining convolutional local feature extraction with self-attention
for global context modeling (rank 4.28). While its accuracy is competitive, this
comes at nearly 50\% higher computational cost compared to MS4N, suggesting diminishing returns relative to its
increased complexity. In addition, MS4N demonstrates approximately 20\% improved stability (lower standard deviation),
further highlighting its robustness compared to competing approaches. At the other extreme, Quant (rank 5.34) highlights that non-deep-learning approaches can remain competitive under strict computational constraints, although its relatively higher variance (0.0291, ~38\% higher than MS4N).

Transformer-based models generally underperform relative to their computational
cost in this setting. The vanilla Transformer operates at 702~MMac yet ranks
only 8.5 overall, while iTransformer ranks 13th with the highest error (0.340).
Informer improves efficiency through ProbSparse attention (rank 7.0), but remains
substantially more expensive than SSM-based models without clear gains in accuracy
or stability. RNNs achieve moderate performance (GRU rank 6th) at higher cost,
while CNN-based models such as HInception—requiring $3\times$ more parameters and
$7\times$ more FLOPs than MS4N—yield higher error and one of the highest STD values.

Overall, the results suggest that well-matched inductive bias and lightweight
design can be more effective than increasing architectural complexity in
TSC. MS4 shows that a minimal structured model
is already competitive, while MS4N demonstrates that a single normalization step
can further improve both accuracy and stability at minimal cost. ConvTran confirms
the effectiveness of hybrid modeling, albeit with higher computational demands,
and Quant illustrates that non-deep simplicity can remain competitive in accuracy
but less consistent in stability. Taken together, these findings favour structured
simplicity with appropriate inductive bias over unconstrained performance,
at least for the continuous TSC tasks evaluated here. In the following, we conduct a domain-level analysis of the MONSTER benchmark. Additional per-dataset results for MONSTER are provided in Table \ref{tab:error_rate_monster}, while domain-wise performance across MONSTER is summarized in Table \ref{tab:domain_model} in the Appendix.

\begin{figure}[htbp]
    \centering
    \includegraphics[width=1\textwidth]{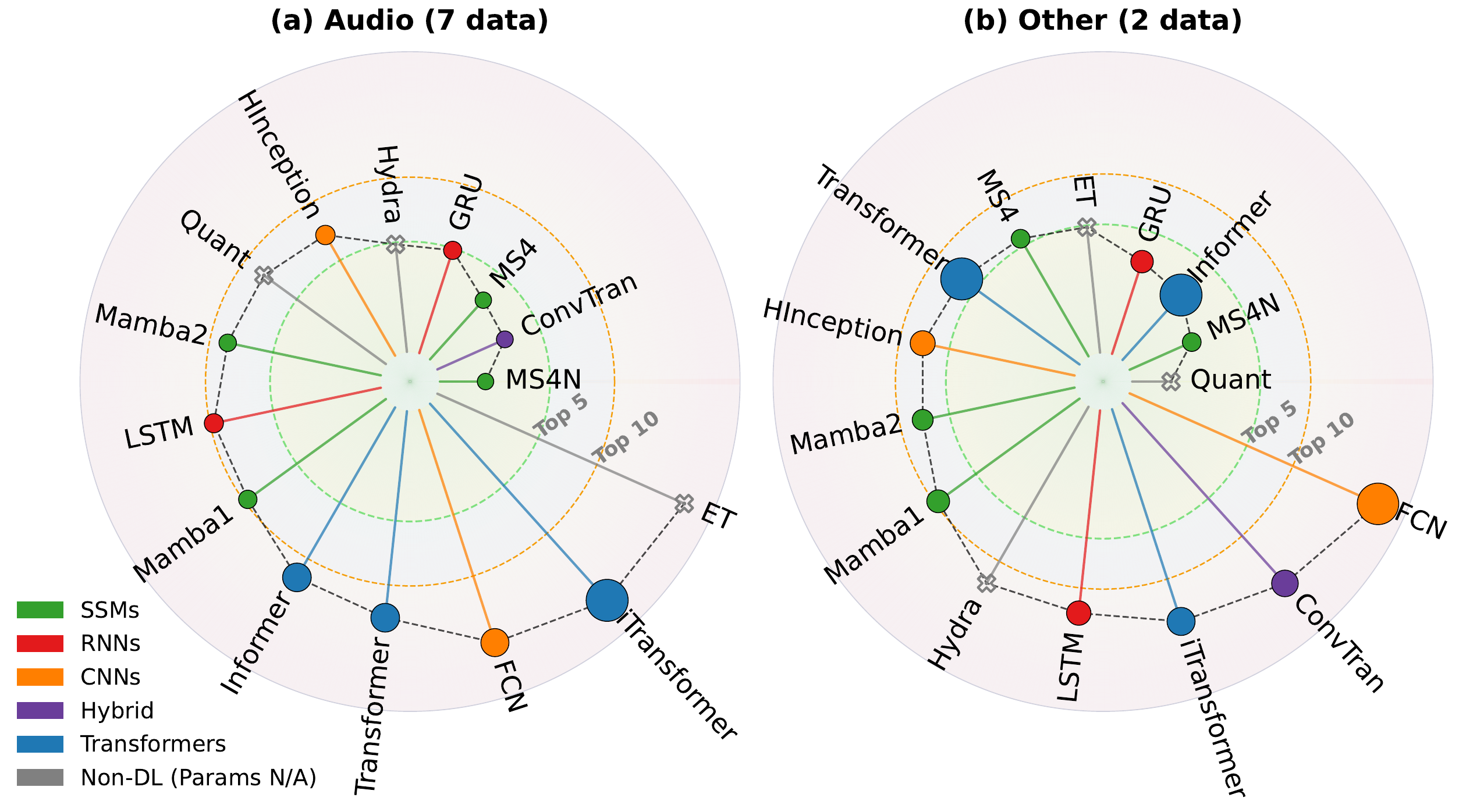}
    \caption{Polar rank comparison of all evaluated models across two domain groups: (a) Audio (7 datasets) and (b) Other (2 datasets), illustrating the complexity--performance trade-off. Each spoke represents a model, with spoke length proportional to accuracy average rank across datasets within the domain (shorter is better). Filled circles indicate deep learning models, with dot size scaled to parameter count, reflecting model complexity; cross ($\times$) markers denote non-deep-learning baselines for which parameter counts are not applicable. Dashed reference rings highlight the Top-5 and Top-10 performance boundaries. The dashed black connector links models in ascending rank order to visualise the performance gap distribution. Models are colour-coded by architecture family: SSMs (green), Transformers (blue), Hybrid (purple), RNNs (red), CNNs (orange), and non-DL baselines (grey).}
    \label{fig:MonsterAudio}
\end{figure}

\subsubsection{Audio domain: long-range temporal dependencies}

On 7 Audio datasets (Figure \ref{fig:MonsterAudio}(a))—whose sequence lengths range from approximately 600 to 50K time steps—model behavior differs noticeably from the aggregate results over all 29 problems. This wide span of temporal resolutions makes the domain particularly suitable for evaluating scalability and long-range modeling capacity.

MS4N achieves the best overall performance with an average rank of 2.28 and an error of 0.10, outperforming the second-best model, ConvTran (rank 3.71), by roughly 2\% in accuracy. The third position is held by MS4 (rank $\approx 4.0$), close to ConvTran, further reinforcing that diagonal S4-based variants are particularly well suited to audio signals—even as sequence lengths extend to tens of thousands of steps.

A notable outcome is the strong performance of GRU (rank 5.42), which surpasses all Transformer-based, CNN-based, and Mamba-based variants in this domain. This suggests that relatively simple recurrent inductive biases remain effective for structured, continuous temporal patterns typical of audio data. However, this comes at a substantial computational cost: GRU requires approximately 500 MMac—nearly six times more than MS4N—making it significantly less efficient despite competitive accuracy, especially for longer sequences.

Among NonDL methods, Hydra (rank 5.42) outperforms Quant (rank 7.57), reversing their relative ordering from the overall benchmark. Within CNN-based models, HInception (rank $\approx 7.0$) is the strongest performer. Mamba variants occupy a middle tier: while not among the top models, they remain more memory-efficient than larger Transformer-based and FCN architectures. The lowest ranks are observed for ET and iTransformer (13.57 and 13.28, respectively).

Overall, across audio datasets spanning 600 to 50K time steps, the results underscore the importance of architectures that scale effectively with sequence length. Lightweight diagonal SSMs maintain both accuracy and efficiency across this broad range.

\subsubsection{Other domain: moderate to large sample sizes}
In the Other domain (Figure~\ref{fig:MonsterAudio}(b)), which comprises only two datasets but spans moderate to large sample sizes (one exceeding one million instances and the other around 36K), MS4N emerges as the best-performing deep learning model (average rank 3.5), positioning it as a top-tier method in this setting. While Quant achieves the overall best rank (2.0, error 0.0626), its advantage reflects the effectiveness of lightweight, feature-based approaches rather than a limitation of data scale.

Among neural architectures, MS4N provides the most favorable balance between predictive performance and computational efficiency, requiring only $\sim$24K parameters and 2.62 FLOPs. In contrast, Transformer-based models (e.g., Informer and Transformer) and Mamba variants incur substantially higher computational cost—often an order of magnitude larger in parameters and FLOPs—yet offer little improvement in ranking or error. This suggests that increasing architectural complexity does not yield proportional gains, even when sufficient data is available.

Similarly, MS4 (rank 7.0, error 0.083) and HInception (rank 8) further illustrate that higher model capacity does not necessarily translate into better performance in these multivariate settings. Overall, the results indicate that structured and computationally efficient models remain highly competitive, even in domains that are not strictly small-scale.

Within this context, MS4N stands out as the most effective deep learning approach, achieving strong accuracy while maintaining a compact and scalable design.

\subsubsection{EEG domain: high-frequency and multivariate}

\begin{figure}[htbp]
    \centering
    \includegraphics[width=1\textwidth]{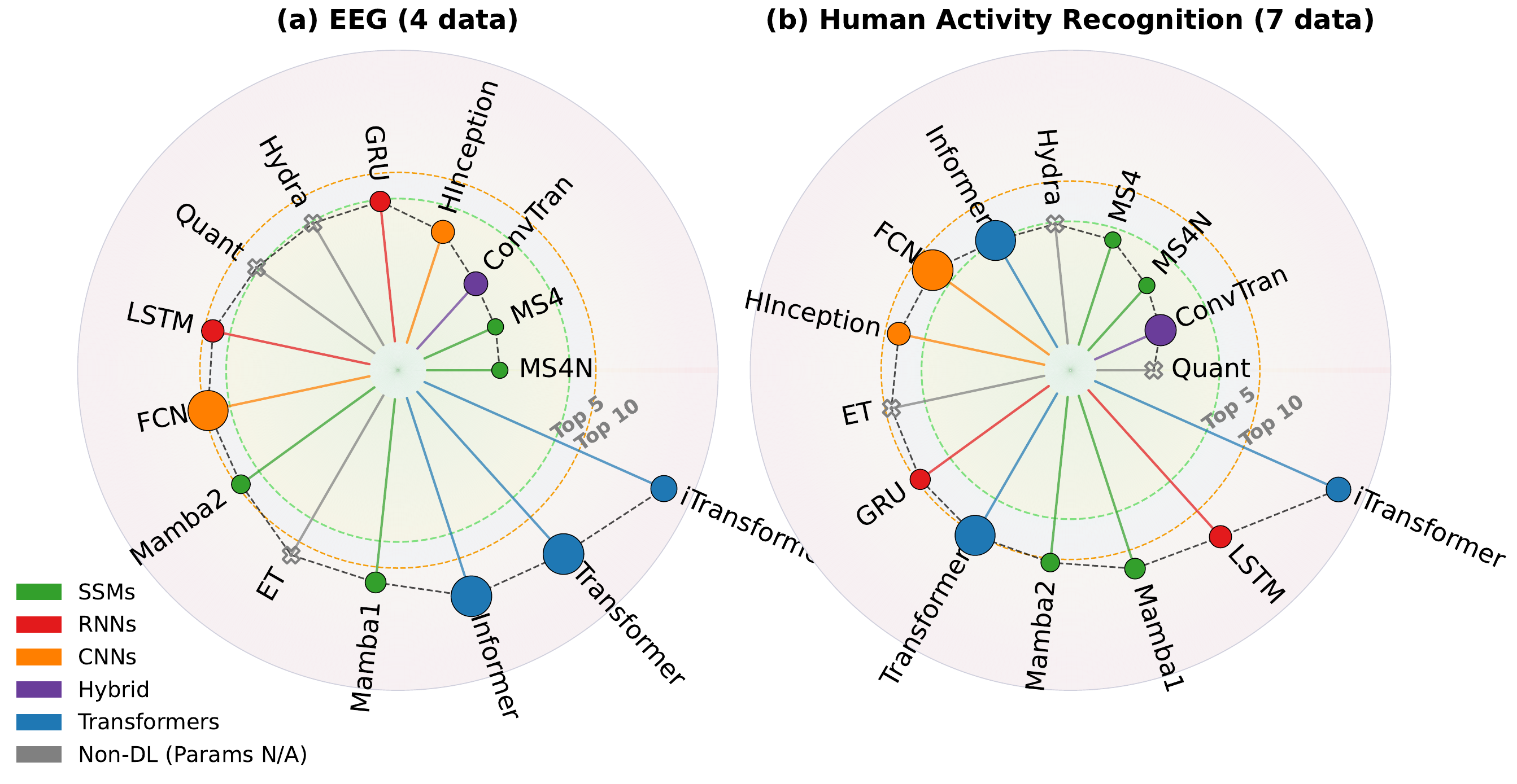}
    \caption{Polar rank comparison of all evaluated models across two domain groups: (a) EEG (4 datasets) and (b) Human Activity Recognition (7 datasets), illustrating the performance (average rank) vs complexity (model size). Each spoke represents a model, with spoke length proportional to accuracy average rank across datasets within the domain (shorter is better). Filled circles indicate deep learning models, with dot size scaled to parameter count, reflecting model complexity; cross ($\times$) markers denote non-deep-learning baselines for which parameter counts are not applicable. Dashed reference rings highlight the Top-5 and Top-10 performance boundaries. The dashed black connector links models in ascending rank order to visualise the performance gap distribution. Models are colour-coded by architecture family.}
    \label{fig:MonsterEEG}
\end{figure}

On the four EEG datasets (Figure \ref{fig:MonsterEEG}(a)), MS4N achieves the best overall performance (rank $3.7$, error rate $0.362$). In addition, MS4 with error rate $0.364$ and rank $4$, outperforms  ConvTran (rank $4.5$, error rate $0.349$), despite maintaining a significantly lighter computational profile.

MS4N remains highly competitive in terms of accuracy while requiring substantially fewer parameters ($\sim 24\text{K}$) and lower FLOPs compared to ConvTran ($\sim 3.4\times$ more parameters and $\sim 11\times$ higher FLOPs). This highlights that structured state-space modeling remains effective even under strict efficiency constraints.

In contrast, ConvTran delivers strong but less efficient performance, indicating diminishing returns from hybrid convolution--attention designs in EEG regimes. Transformer-based methods such as iTransformer and Informer perform significantly worse (ranks $11.0$--$13.5$, respectively), reinforcing their limited data efficiency and weak inductive bias for neural time-series. Intermediate baselines such as HInception (rank $6.0$) and GRU (rank $7.0$) show moderate accuracy--efficiency trade-offs and are consistently outperformed by structured state-space and hybrid architectures.

Overall, the results demonstrate that MS4N provides the strongest accuracy--efficiency balance in EEG modeling, suggesting that structured temporal inductive bias is more critical than scale or architectural complexity in low-data neural signal regimes.

\subsubsection{Human activity recognition (HAR)}

The HAR domain (Figure~\ref{fig:MonsterEEG} b) highlights the importance of efficient cross-channel modeling in high-dimensional time series, where sensor counts can exceed 100 channels. In this setting, architectures relying on quadratic channel interactions, such as Transformers, incur substantial computational overhead.

MS4N achieves a top-tier performance, while requiring only 25K parameters and 1.14 MMac—representing a 10–25× reduction in computational cost compared to Transformer-based models such as ConvTran and Informer. Notably, ConvTran attains comparable accuracy (rank 4.0) but at over 20× higher FLOPs.

Importantly, even the simpler MS4 baseline demonstrates consistently stronger performance than Mamba-based models (Mamba1 and Mamba2), while operating at significantly lower computational cost. This suggests that explicit and efficient channel mixing is more effective than input-dependent state dynamics in high-dimensional settings, where stable and structured representations are critical.

Overall, these results indicate that structured state space models with lightweight channel interaction mechanisms can effectively capture high-dimensional dependencies without resorting to costly quadratic or input-dependent operations.
\subsubsection{Satellite image time series}

On the seven Image Satellite Time Series (ISTS) datasets (Figure \ref{fig:MonsterCounts}(a))—some comprising up to 60 million samples—the ranking differs notably from the Audio domain. In this regime, computational cost, memory footprint, and scalability become critical evaluation factors.

ConvTran achieves the best overall performance (error rate $0.1511$, rank $2.86$), followed by Informer (rank $4.0$) and the vanilla Transformer (rank $5.14$). However, these Transformer-based models incur substantially higher computational and memory costs, with approximately $287\text{K}$ parameters and $26$ FLOPs on average. Their strong performance in this domain is likely enabled by the availability of large-scale training data, which compensates for their relatively weak inductive bias.

In contrast, MS4N (rank $8.43$, error rate $0.1787$) achieves competitive performance while remaining significantly more efficient, requiring only $\sim 24.7\text{K}$ parameters and $\sim 0.85$ FLOPs—representing a $10$--$12\times$ reduction in parameters and approximately $30\times$ lower computational cost compared to Informer and Transformer models. MS4 (rank $8.0$) exhibits a similarly lightweight profile. Non-deep learning methods such as Quant (rank $5.71$) and ET (rank $7.86$) also demonstrate strong efficiency, while recurrent models (GRU, LSTM) and Mamba variants provide moderate trade-offs without achieving top performance.

These results highlight an important distinction between model classes. ISTS datasets are characterized by large sample sizes and relatively structured temporal patterns, where the abundance of data reduces reliance on strong inductive biases. In such settings, the flexibility of self-attention enables Transformer-based models to effectively capture global dependencies.

Overall, while Transformer-based models achieve the lowest error rates on ISTS, MS4N provides a substantially better efficiency–performance trade-off, making it a practical and scalable choice for large-scale satellite time-series applications.

\subsubsection{Counts domain}

\begin{figure}[htbp]
    \centering
    \includegraphics[width=1\textwidth]{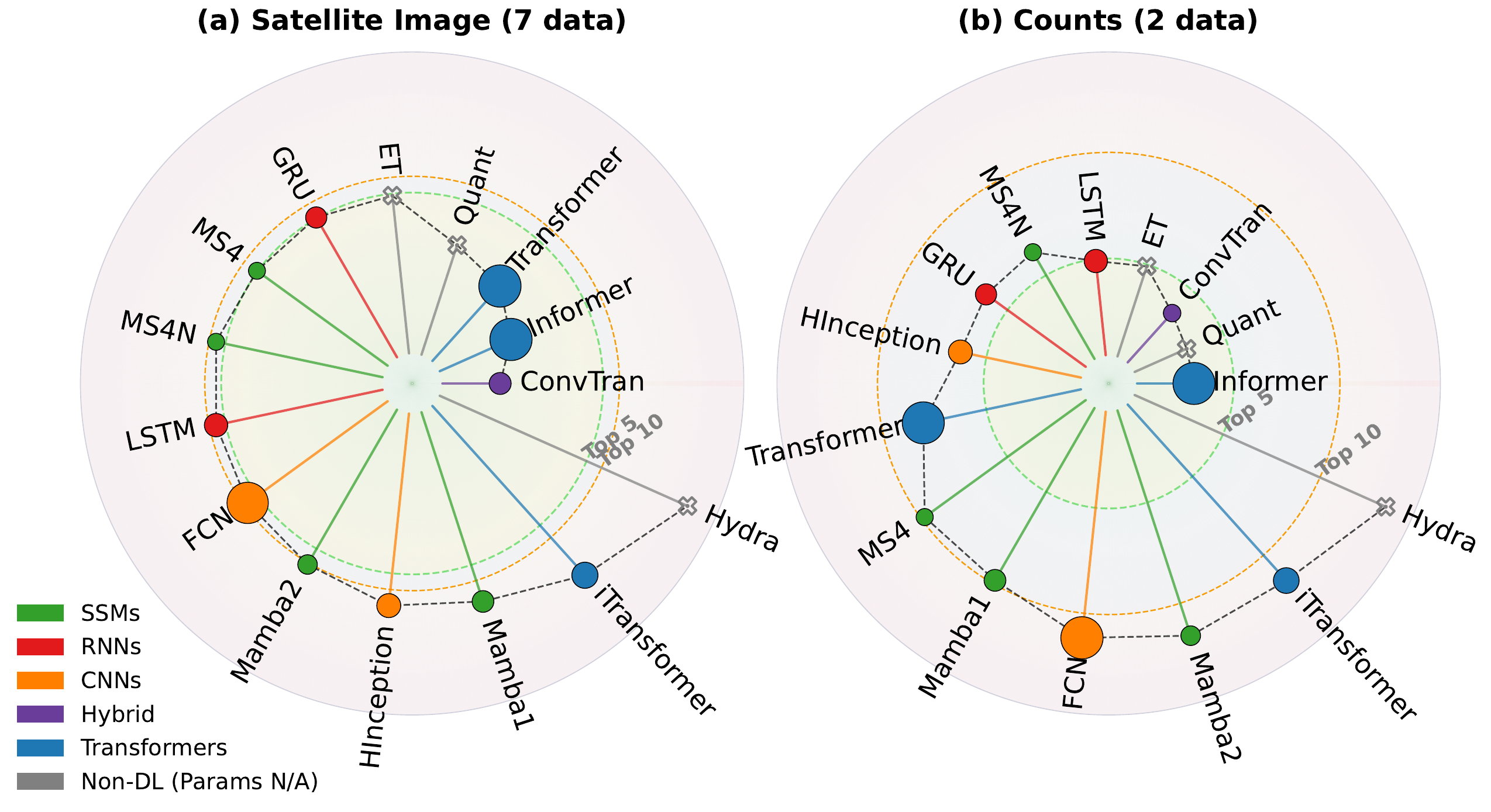}
    \caption{Polar rank comparison of all evaluated models across two domain groups: (a) Satellite Image Time Series (7 datasets) and (b) Counts (2 datasets). Each spoke represents a model, with spoke length proportional to its accuracy average rank across datasets within the domain (shorter is better), and dot size scaled to parameter count for deep learning models, jointly visualising the performance--complexity trade-off.}
    \label{fig:MonsterCounts}
\end{figure}

The Counts (univariate, short sequences) (Figure \ref{fig:MonsterCounts}(b))  domain highlight scenarios where simplicity, efficiency, and appropriate model capacity dominate over raw representational power. 

In the Counts domain, Quant achieves the lowest error (0.2299) and the best average rank (3.0), demonstrating that lightweight models with minimal parameters perform best on short univariate sequences. Deep models such as ConvTran and Informer achieve similar ranking in some datasets, but at a far higher computational cost (ConvTran: 28K parameters, 0.70 FLOPs; Informer: 287K parameters, 6.82 FLOPs). This contrast emphasizes that for short sequences, overparameterized attention-based models are often overcapacity, offering marginal gains in accuracy while significantly increasing memory and computation requirements. SSM-based models like MS4N remain efficient but are less competitive in raw accuracy, highlighting the trade-off between structural inductive bias and sequence simplicity.

Overall, these results suggest that Quant performs well in the Counts domain, particularly on short, univariate sequences, providing a good balance of accuracy and efficiency. MS4N and MS4 offer a reasonable trade-off in the Counts domain, and appear particularly efficient.

\medskip
\textbf{Summary:} Across the MONSTER benchmark, MS4N and MS4 achieve favorable efficiency–accuracy trade-offs, generally outperforming larger Transformers, RNNs, and CNNs across most domains. Hybrid models like ConvTran also show promising performance in several domains but with high complexity cost. Among non-deep learning models, Quant achieves relatively high accuracy while maintaining good efficiency. Attention-based models perform well mainly on ISTS, whereas NonDL approaches remain competitive on short or analytically tractable sequences. Overall, alignment between model architecture and the temporal or cross-channel structure of the data appears to be a more important factor for performance than raw parameter count.

\subsection{Experiments on UEA benchmark: multivariate TSC}
\begin{figure}[htbp]
    \centering
    \includegraphics[width=1\textwidth]{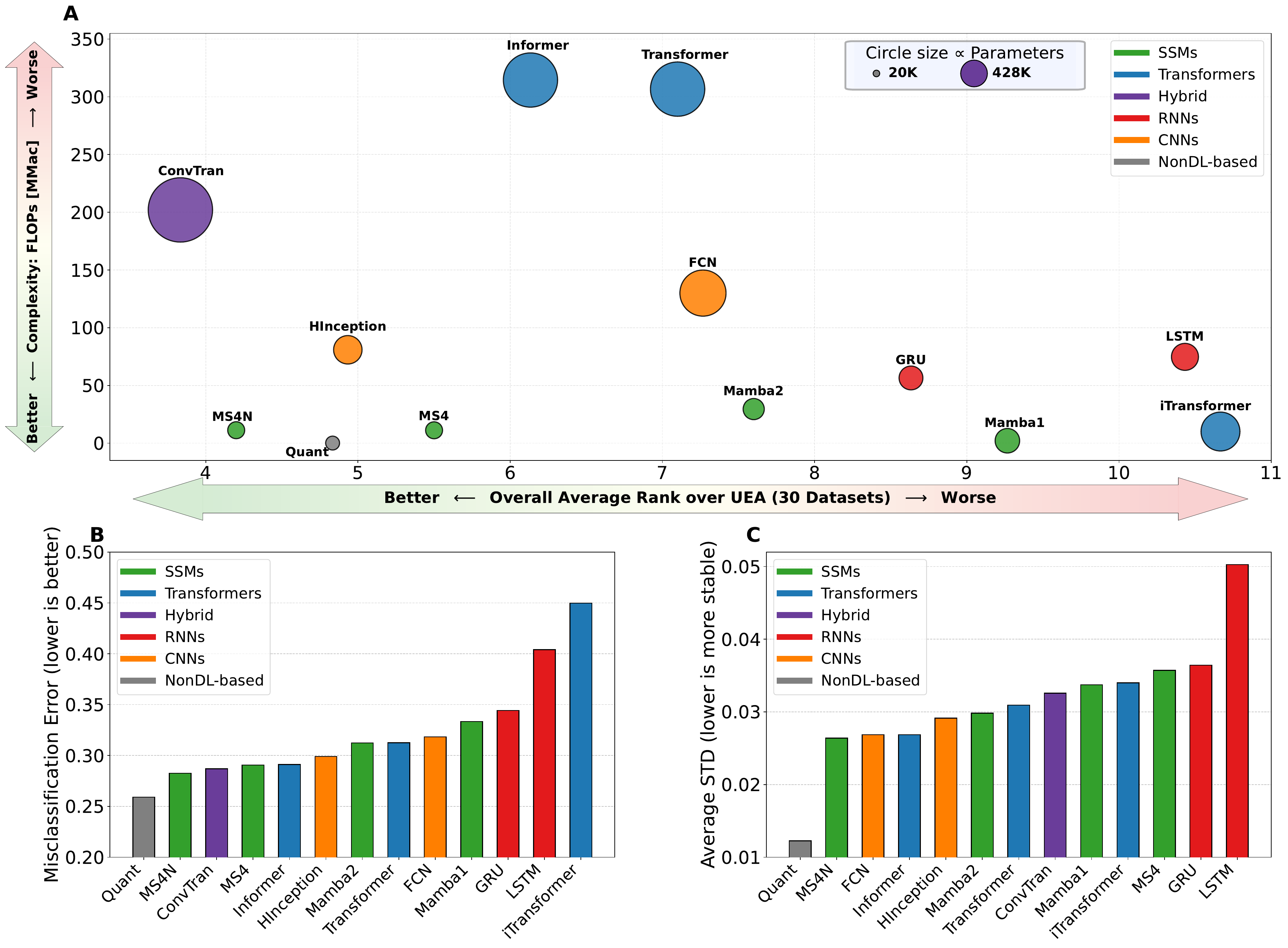}
    \caption{Three-panel comparison of all models over UEA multivariate benchmark including 30 datasets. (A) Bubble plot of
overall average rank versus computational cost (FLOPs in MMac); bubble size is proportional to
the number of parameters—models in the bottom-left are both accurate and efficient. (B) Average
misclassification error per model, sorted in ascending order. (C) Average standard deviation (STD)
per model, sorted in ascending order—lower values indicate more stable training. All panels are
colored by architectural family.}
    \label{fig:UEA}
\end{figure}

Figure~\ref{fig:UEA} summarizes the performance of 13 models evaluated on the UEA benchmark across 30 multivariate time series classification (MTSC) datasets, reporting average error rate, rank, number of parameters, and FLOPs. Among all deep learning models in the comparison, MS4N achieves the lowest average error rate of 0.2826, demonstrating superior classification performance on this benchmark. While MS4N ranks second in average rank (4.2) behind ConvTran (3.83), this ranking advantage comes at a substantial computational cost: ConvTran requires 428,189 parameters and 202.04 FLOPs, compared to MS4N's compact profile of only 29,920 parameters and 11.24 FLOPs — making ConvTran approximately 14× larger in parameters and 18× more expensive in FLOPs than MS4N. This positions MS4N as the most efficient deep learning model in the comparison, achieving the lowest error rate at a fraction of the budget required by its closest competitor. To provide a broader point of reference, we also include QUANT, the state-of-the-art non-deep learning model for MTSC. QUANT ranks third overall with an average rank of 4.8, making it the strongest non-deep learning baseline in terms of both rank and error rate, further contextualizing the competitiveness of MS4N across the full model spectrum.

The contribution of the normalization layer is further evidenced by comparing MS4N with its predecessor MS4, which shares an identical architecture but omits normalization. MS4 yields a higher average error rate of 0.2907 and an average rank of 4.93, whereas MS4N reduces both 
metrics while introducing negligible overhead, as the two models share nearly identical parameter counts and FLOPs. This confirms that the normalization layer provides a consistent and meaningful performance gain without increasing model complexity.

Among CNN-based models, HInception achieves a competitive error rate of 0.2990, outperforming FCN (0.3184) and all other CNN variants. However, this comes at a cost of 83,665 parameters and 80.89 FLOPs, making it considerably heavier than MS4N.

Among Transformer-based models, Informer and the standard Transformer yield error rates of 0.2911 and 0.3125, ranking fifth and sixth overall, respectively. iTransformer performs the weakest within this family with an error rate of 0.4497, suggesting that generic attention mechanisms may require task-specific adaptations to handle the diversity of TSC tasks present in the UEA benchmark. Additionally, Informer 
incurs the highest computational cost across all evaluated models at 314.53 MFLOPs, making it the least efficient Transformer variant in this comparison.

Among SSM variants, Mamba1 and Mamba2 achieve error rates of 0.3333 and 0.3124, respectively, placing them in the mid-range of the comparison despite their relatively moderate parameter counts of 62,176 and 46,352. 
Recurrent architectures such as GRU and LSTM rank among the weakest performers, with error rates of 0.3442 and 0.4040, reflecting well-documented limitations of sequential models in capturing complex multivariate temporal dependencies.

Overall, these results highlight that MS4N achieves the best trade-off between accuracy, rank, and computational efficiency across the UEA benchmark, consistently outperforming CNN-based, Transformer-based, recurrent, and state-space model families alike. While ConvTran demonstrates competitive effectiveness by achieving the best average rank, it is 
approximately 14$\times$ larger in parameters and 18$\times$ more expensive in FLOPs than MS4N, making it a prohibitively costly alternative under resource-constrained settings. Furthermore, the improvement of MS4N over MS4 is consistent with earlier observations on the MONSTER benchmark and reinforces the effectiveness of incorporating a normalization layer, which contributes to enhanced accuracy and robustness. Additional details for UEA benchmark provided in Table \ref{tab:error_rate_uea} in the Appendix.

\subsection{Ablation study}

The integration of a normalization layer improves the MS4 architecture in many cases, yielding gains in convergence speed, classification accuracy, and robustness, although the magnitude of improvement varies across datasets. These gains are achieved with minimal computational overhead, 
adding only $N$ parameters — representing less than 0.3\% of the total model size.

To evaluate these improvements in detail, we selected two representative subsets 
from the Counts domain: Pedestrian, the most complex subset 
featuring the maximum number of classes (82) and Traffic, characterized by 
high-volume sample data, to demonstrate the effect of normalization on training 
convergence speed.

\begin{figure}[htbp]
    \centering
    \includegraphics[width=0.85\textwidth]{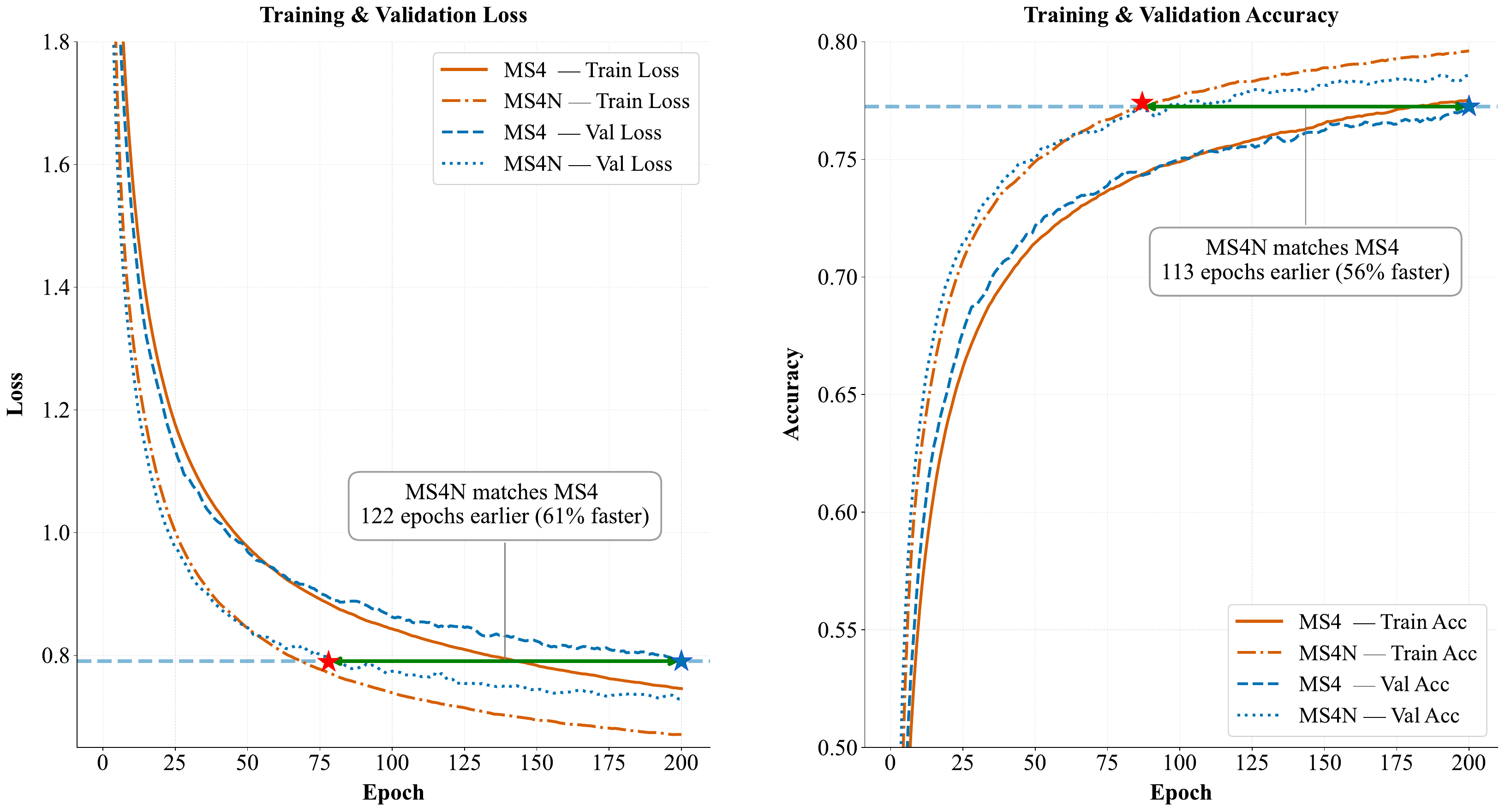}
    \caption{Convergence comparison between MS4 and MS4N on the Pedestrian dataset 
    over 200 epochs. Left: Training and validation loss curves. Right: Training and 
    validation accuracy curves. Stars mark the epoch at which each model reaches 
    the reference threshold (dashed cyan line).}
    \label{fig:convergence_ped}
\end{figure}

Figure~\ref{fig:convergence_ped} presents the training and validation dynamics on 
the Pedestrian dataset. In the loss curves (left panel), MS4N's validation 
loss crosses the reference threshold at epoch 78, while MS4 only reaches the same 
threshold at epoch 200 — a gap of 122 epochs, making MS4N 61\% faster 
than its unnormalized counterpart on this data. A similar pattern emerges in the accuracy 
curves (right panel), where MS4N reaches the reference target at epoch 74 compared 
to epoch 200 for MS4, corresponding to a convergence advantage of 126 epochs, or 
63\% faster. MS4N also attains a higher peak training accuracy, indicating 
that the normalization layer not only accelerates convergence but also enhances the 
model's overall learning capacity under high class diversity.

\begin{figure}[htbp]
    \centering
    \includegraphics[width=0.85\textwidth]{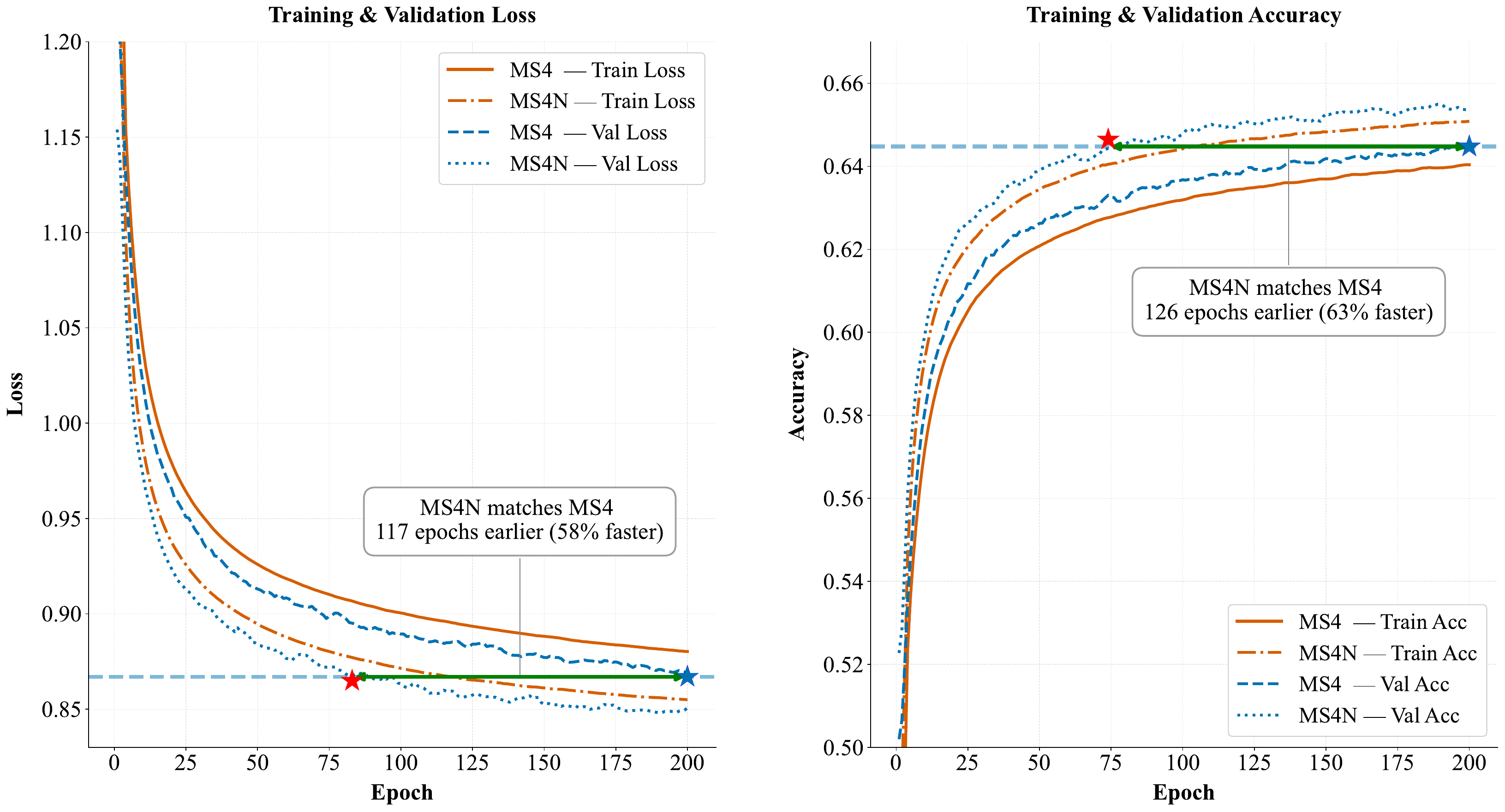}
    \caption{Convergence comparison between MS4 and MS4N on the Traffic dataset 
    over 200 epochs. Left: Training and validation loss curves. Right: Training and 
    validation accuracy curves. Stars mark the epoch at which each model reaches 
    the reference threshold (dashed cyan line).}
    \label{fig:convergence_traffic}
\end{figure}

Figure~\ref{fig:convergence_traffic} presents the corresponding dynamics on the 
Traffic dataset. In the loss curves 
(left panel), MS4N's validation loss crosses the reference threshold at epoch 83, 
while MS4 requires the full 200 epochs — a gap of 117 epochs, corresponding to a 
58\% reduction in training time. Similarly, in the accuracy curves (right 
panel), MS4N reaches the reference validation accuracy target at epoch 74, compared 
to epoch 200 for MS4, yielding a convergence advantage of 126 epochs, or 
63\% faster. MS4N also attains a higher peak training accuracy, 
reinforcing the conclusion that the normalization layer consistently accelerates 
convergence and improves learning efficiency across datasets of varying complexity 
and scale.

Taken together, these results show that the normalization layer is a critical 
architectural component of MS4N. Across both the Pedestrian and Traffic subsets, 
MS4N converges significantly faster — by over 100 epochs (more than twice) in both cases — while 
achieving superior or equivalent final performance. This consistent behavior across 
datasets of different characteristics underscores the generalizability and 
practical value of the proposed normalization mechanism.

\begin{figure}[htbp]
    \centering
    \includegraphics[width=1\textwidth]{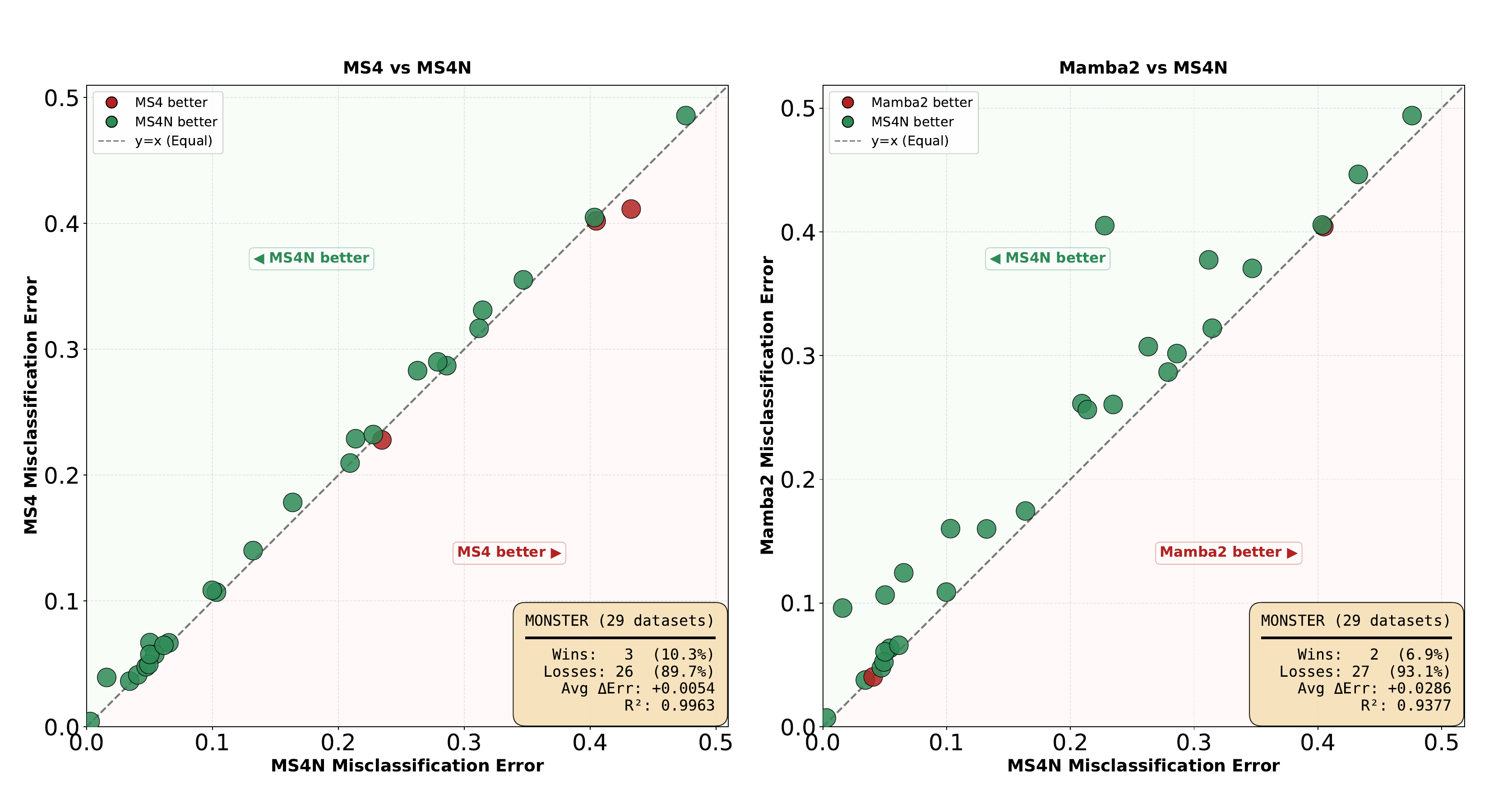}
    \caption{ Comparison of MS4 and Mamba2 against MS4N across 29 benchmark datasets. Each point represents a single dataset. The dashed diagonal line indicates equal performance; points in the upper-left region (green shading) indicate MS4N superiority, while points in the lower-right region (red shading) indicate the competing model's superiority. }
    \label{fig:MS4N-MS4-Mamba2}
\end{figure}

To further ablate the contribution of the architectural modifications introduced 
in MS4N, Figure~\ref{fig:MS4N-MS4-Mamba2} compares its misclassification error 
against both its direct predecessor MS4 and the Mamba2 backbone across 29 MONSTER benchmark 
datasets. In both ablation settings, the vast majority of points lie above the 
diagonal, indicating that MS4N consistently achieves lower misclassification error. 
Against MS4, this ablation reveals that the normalization layer yields measurable 
gains on 27 out of 29 datasets, suggesting that improvements are 
uniform regardless of dataset difficulty. The ablation against Mamba2 further 
confirms this trend, with MS4N outperforming on 26 out of 29 datasets, reflecting a more pronounced 
performance gap at the backbone level — noteworthy given that Mamba2 requires 
twice the parameters and FLOPs of MS4N. Taken together, these ablation results 
demonstrate that the design choices in MS4N generalize robustly across diverse 
TSC benchmarks.

Table~\ref{tab:robustnessSSM} reports the average standard deviation of performance across all MONSTER datasets for different variants within the SSM family, serving as a measure of robustness. Lower standard deviation indicates more consistent behavior under varying conditions.

Within this family, MS4N achieves the lowest standard deviation, suggesting that the proposed normalization contributes to more stable and reliable performance. MS4 shows slightly higher variability, while Mamba2 and Mamba1 exhibit progressively larger fluctuations, indicating reduced stability. These results provide an ablation-style insight, showing that the incremental design choices—from Mamba variants to MS4 and further to MS4N—lead to improved robustness, with normalization playing a key role in stabilizing model behavior across different settings.

\begin{table}[t]
\centering
\caption{Robustness comparison in SSM-based models}
\label{tab:robustnessSSM}
\begin{tabular}{lcccc}
\toprule
Model & MS4N & MS4 & Mamba2 & Mamba1 \\
\midrule
Average STD & \textbf{0.0212} & 0.0246 & 0.0272 & 0.0373 \\
\bottomrule
\end{tabular}
\end{table}
\subsubsection{Ablation on increasing layers and hidden size}

To investigate whether added capacity translates into improved time series
classification performance, we ablate MS4 along two dimensions on the MONSTER
benchmark: the number of stacked layers and the hidden dimension size.
Table~\ref{tab:MS4variant} summarizes the performance--complexity trade-offs
across variants. Specifically, it reports the overall average rank across all
29 MONSTER datasets, alongside computational cost (in MMac) and the number of
trainable parameters. This comparison allows us to directly assess whether
increased model capacity yields consistent improvements in ranking performance
relative to its computational overhead.

\begin{table}[h]
\centering
\caption{Ablation of MS4 variants on MONSTER}
\label{tab:MS4variant}
\begin{tabular}{lcccc}
\hline
Model & Error & Rank & Parameters & FLOPs \\
\hline
MS4N & \textbf{0.1872} & \textbf{1.72} & 24676 & 21.53 \\
MS4-2Layer & 0.1899 & 2.48 & 43300 & 52.08 \\
MS4-H128 & 0.1920 & 2.83 & 85948 & 82.73 \\
MS4 & 0.1907 & 2.97 & \textbf{24548} & \textbf{21.37} \\
\hline
\end{tabular}
\end{table}
Both MS4-2Layer and MS4-H128 scaling strategies improve classification rank over the MS4 baseline, yet at
a steep computational cost. Doubling the hidden dimension (MS4-H128) raises FLOPs
by roughly $4\times$ and expands the parameter count to ${\sim}86$K, while
stacking a second layer (MS4-2Layer) achieves a better rank-complexity trade-off
than MS4-H128---indicating that depth is a more efficient route to classification
accuracy than width across the MONSTER datasets. Nevertheless, both variants
remain substantially more expensive than necessary.

MS4N (1 layer, $H{=}64$, ${\sim}25$K parameters) achieves the best average rank
across all 29 MONSTER classification tasks while matching MS4 in parameter count
and computational cost. It outperforms even the much largerand costlier MS4-H128 and MS4-2Layer, demonstrating that targeted architecturaldesign---rather than brute-force capacity---is the key driver of classification performance. These results reinforce a clear message: \emph{complexity does not imply competence}, and MS4N represents the optimal point in the
accuracy--efficiency trade-off for TSC.

\section{Conclusion}

In this work, we presented a systematic and comprehensive study of the structured SSM family for TSC, investigating whether architectural advances within this family—most notably the shift toward input-dependent dynamics in Mamba—yield consistent benefits over simpler diagonal alternatives. Building on this foundation, we revisited diagonal structured state-space models (S4D) as a simple yet effective foundation for TSC, providing an alternative perspective to the prevailing trend toward increasingly complex architectures, and introduced MS4, which augments S4D with a learnable input projection and a lightweight channel-mixing module, and MS4N, a normalized variant that incorporates a single LayerNorm operation to improve training stability, convergence, and robustness with negligible computational overhead.

To rigorously assess model suitability, we adopt a multi-criteria evaluation framework encompassing classification accuracy, computational complexity, model size, memory efficiency, and robustness. Extensive empirical evaluations across the MONSTER and UEA benchmarks demonstrate that MS4 serves as a strong and efficient baseline, while MS4N further improves accuracy, achieving competitive—and in many cases state-of-the-art—results across a diverse set of 15 representative models. Notably, MS4N consistently delivers strong predictive accuracy while requiring significantly fewer parameters and lower computational cost than Transformer-, Mamba-, CNN-, and RNN-based approaches.  These advantages are particularly evident in domains such as Audio, EEG, and Human Activity Recognition, where long sequences, noise, and high dimensionality favor structured temporal modeling.

At the same time, our results indicate that no single architecture is universally optimal. MS4N demonstrates that structured, input-invariant dynamics provide a strong inductive bias, improving robustness and efficiency in challenging regimes, including long, high-frequency, and high-dimensional time-series data. In contrast, hybrid models such as ConvTran remain highly competitive in more structured settings, but their computational complexity scales unfavorably with increasing channel dimensionality, limiting their applicability in high-dimensional multivariate scenarios. Transformer-based approaches, such as Informer, can achieve lower error rates in large-scale settings where abundant data allows their flexibility to be fully exploited. These observations highlight the importance of aligning model architecture with data characteristics, rather than relying solely on increasing model complexity.

Overall, this work shows that carefully designed, lightweight structured models can match or surpass substantially more complex architectures, offering a compelling and scalable alternative for TSC. Future work will explore extending MS4N to forecasting and anomaly detection tasks, investigating adaptive channel-mixing mechanisms, and developing a deeper theoretical understanding of structured state-space models, particularly in real-time and resource-constrained settings.

\bibliography{Ref}
\appendix

\section{Additional Results}

Table~\ref{tab:error_rate_monster} reports the misclassification error and standard deviation 
for all models across all 29 datasets in MONSTER, with results averaged over five-fold 
cross-validation. 

Table~\ref{tab:domain_model} provides a domain-level breakdown of MONSTER results across 
six categories: Audio (7 datasets), Satellite Imagery (7 datasets), Human Activity 
Recognition (7 datasets), EEG (4 datasets), Counts (2 datasets), and Other (2 datasets). 
For each domain, models are evaluated along four criteria --- misclassification error, average 
rank, number of parameters, and FLOPs --- offering a comprehensive, multi-faceted comparison 
across all models and domains.

Table~\ref{tab:error_rate_uea} presents the corresponding results across 
all 30 datasets of the UEA benchmark, averaged over 40 independent runs.

\begin{sidewaystable}[htbp]
\centering
\caption{Error rate ${\pm}$ STD per dataset and model over MONSTER}
\label{tab:error_rate_monster}
\tiny
\fontsize{6}{7}\selectfont
\setlength{\tabcolsep}{3pt}

\begin{tabular}{lrrrrrrr}
\toprule
Dataset & MS4N & MS4 & Mamba1 & Mamba2 & ConvTran & HInception & FCN \\
\midrule
AudioMNIST            & $0.018{\pm}0.007$ & $0.039{\pm}0.018$ & $0.128{\pm}0.028$ & $0.096{\pm}0.033$ & $0.014{\pm}0.006$ & $0.211{\pm}0.028$ & $0.896{\pm}0.009$ \\
AudioMNIST-DS         & $0.048{\pm}0.013$ & $0.067{\pm}0.020$ & $0.138{\pm}0.022$ & $0.106{\pm}0.020$ & $0.051{\pm}0.013$ & $0.074{\pm}0.009$ & $0.880{\pm}0.023$ \\
CornellWhaleChallenge & $0.105{\pm}0.004$ & $0.107{\pm}0.006$ & $0.156{\pm}0.008$ & $0.160{\pm}0.011$ & $0.090{\pm}0.005$ & $0.155{\pm}0.005$ & $0.160{\pm}0.014$ \\
CrowdSourced          & $0.324{\pm}0.043$ & $0.331{\pm}0.054$ & $0.332{\pm}0.121$ & $0.322{\pm}0.105$ & $0.265{\pm}0.114$ & $0.334{\pm}0.124$ & $0.346{\pm}0.123$ \\
DREAMERA              & $0.475{\pm}0.035$ & $0.486{\pm}0.026$ & $0.487{\pm}0.032$ & $0.494{\pm}0.028$ & $0.476{\pm}0.058$ & $0.529{\pm}0.045$ & $0.486{\pm}0.064$ \\
DREAMERV              & $0.432{\pm}0.022$ & $0.411{\pm}0.035$ & $0.443{\pm}0.034$ & $0.446{\pm}0.021$ & $0.437{\pm}0.030$ & $0.388{\pm}0.017$ & $0.421{\pm}0.034$ \\
FordChallenge         & $0.100{\pm}0.011$ & $0.108{\pm}0.014$ & $0.106{\pm}0.009$ & $0.109{\pm}0.013$ & $0.139{\pm}0.033$ & $0.099{\pm}0.016$ & $0.188{\pm}0.100$ \\
FruitFlies            & $0.031{\pm}0.002$ & $0.036{\pm}0.002$ & $0.055{\pm}0.026$ & $0.038{\pm}0.004$ & $0.029{\pm}0.003$ & $0.033{\pm}0.005$ & $0.478{\pm}0.191$ \\
InsectSound           & $0.209{\pm}0.007$ & $0.210{\pm}0.003$ & $0.250{\pm}0.006$ & $0.261{\pm}0.005$ & $0.203{\pm}0.007$ & $0.182{\pm}0.007$ & $0.214{\pm}0.013$ \\
LakeIce               & $0.003{\pm}0.001$ & $0.004{\pm}0.001$ & $0.006{\pm}0.001$ & $0.007{\pm}0.001$ & $0.003{\pm}0.001$ & $0.005{\pm}0.001$ & $0.004{\pm}0.000$ \\
LenDB                 & $0.055{\pm}0.012$ & $0.057{\pm}0.012$ & $0.177{\pm}0.220$ & $0.063{\pm}0.011$ & $0.488{\pm}0.033$ & $0.452{\pm}0.046$ & $0.512{\pm}0.028$ \\
MosquitoSound         & $0.065{\pm}0.000$ & $0.067{\pm}0.002$ & $0.112{\pm}0.002$ & $0.124{\pm}0.003$ & $0.193{\pm}0.048$ & $0.170{\pm}0.017$ & $0.592{\pm}0.091$ \\
Opportunity           & $0.163{\pm}0.018$ & $0.178{\pm}0.016$ & $0.172{\pm}0.021$ & $0.174{\pm}0.026$ & $0.164{\pm}0.013$ & $0.175{\pm}0.018$ & $0.163{\pm}0.020$ \\
PAMAP2                & $0.292{\pm}0.102$ & $0.287{\pm}0.140$ & $0.275{\pm}0.114$ & $0.302{\pm}0.138$ & $0.230{\pm}0.080$ & $0.340{\pm}0.287$ & $0.293{\pm}0.177$ \\
Pedestrian            & $0.214{\pm}0.002$ & $0.229{\pm}0.003$ & $0.247{\pm}0.002$ & $0.256{\pm}0.003$ & $0.201{\pm}0.002$ & $0.215{\pm}0.017$ & $0.487{\pm}0.034$ \\
S2Agri-10pc-17        & $0.408{\pm}0.010$ & $0.402{\pm}0.014$ & $0.406{\pm}0.010$ & $0.404{\pm}0.009$ & $0.339{\pm}0.008$ & $0.407{\pm}0.010$ & $0.407{\pm}0.010$ \\
S2Agri-10pc-34        & $0.405{\pm}0.011$ & $0.405{\pm}0.008$ & $0.406{\pm}0.010$ & $0.406{\pm}0.011$ & $0.364{\pm}0.005$ & $0.407{\pm}0.010$ & $0.407{\pm}0.010$ \\
S2Agri-17             & $0.040{\pm}0.001$ & $0.041{\pm}0.002$ & $0.044{\pm}0.001$ & $0.040{\pm}0.001$ & $0.037{\pm}0.001$ & $0.038{\pm}0.001$ & $0.039{\pm}0.001$ \\
S2Agri-34             & $0.047{\pm}0.002$ & $0.048{\pm}0.002$ & $0.051{\pm}0.001$ & $0.048{\pm}0.001$ & $0.044{\pm}0.001$ & $0.047{\pm}0.002$ & $0.049{\pm}0.003$ \\
STEW                  & $0.234{\pm}0.014$ & $0.228{\pm}0.023$ & $0.269{\pm}0.024$ & $0.261{\pm}0.020$ & $0.219{\pm}0.005$ & $0.216{\pm}0.004$ & $0.263{\pm}0.021$ \\
Skoda                 & $0.049{\pm}0.003$ & $0.050{\pm}0.002$ & $0.059{\pm}0.005$ & $0.052{\pm}0.003$ & $0.057{\pm}0.031$ & $0.056{\pm}0.005$ & $0.058{\pm}0.015$ \\
TimeSen2Crop          & $0.047{\pm}0.017$ & $0.057{\pm}0.024$ & $0.059{\pm}0.019$ & $0.061{\pm}0.015$ & $0.039{\pm}0.014$ & $0.093{\pm}0.022$ & $0.136{\pm}0.049$ \\
Tiselac               & $0.298{\pm}0.053$ & $0.290{\pm}0.055$ & $0.287{\pm}0.059$ & $0.287{\pm}0.050$ & $0.232{\pm}0.038$ & $0.292{\pm}0.034$ & $0.239{\pm}0.028$ \\
Traffic               & $0.347{\pm}0.003$ & $0.355{\pm}0.001$ & $0.355{\pm}0.002$ & $0.371{\pm}0.001$ & $0.330{\pm}0.002$ & $0.333{\pm}0.001$ & $0.350{\pm}0.003$ \\
UCIActivity           & $0.050{\pm}0.033$ & $0.065{\pm}0.044$ & $0.073{\pm}0.046$ & $0.066{\pm}0.042$ & $0.059{\pm}0.034$ & $0.050{\pm}0.037$ & $0.059{\pm}0.035$ \\
USCActivity           & $0.320{\pm}0.060$ & $0.317{\pm}0.068$ & $0.357{\pm}0.086$ & $0.377{\pm}0.059$ & $0.291{\pm}0.052$ & $0.313{\pm}0.087$ & $0.338{\pm}0.067$ \\
WISDM                 & $0.150{\pm}0.047$ & $0.140{\pm}0.046$ & $0.169{\pm}0.056$ & $0.160{\pm}0.086$ & $0.141{\pm}0.061$ & $0.188{\pm}0.064$ & $0.167{\pm}0.056$ \\
WISDM2                & $0.284{\pm}0.075$ & $0.283{\pm}0.068$ & $0.316{\pm}0.110$ & $0.307{\pm}0.062$ & $0.258{\pm}0.004$ & $0.293{\pm}0.020$ & $0.288{\pm}0.028$ \\
WhaleSounds           & $0.227{\pm}0.002$ & $0.232{\pm}0.002$ & $0.417{\pm}0.010$ & $0.405{\pm}0.005$ & $0.270{\pm}0.042$ & $0.485{\pm}0.074$ & $0.533{\pm}0.052$ \\
\bottomrule
\end{tabular}

\vspace{4pt}

\begin{tabular}{lrrrrrrr}
\toprule
Dataset & Transformer & Informer & GRU & LSTM & Quant & Hydra & ET \\
\midrule
AudioMNIST            & $0.191{\pm}0.010$ & $0.175{\pm}0.021$ & $0.900{\pm}0.010$ & $0.900{\pm}0.010$ & $0.121{\pm}0.039$ & $0.027{\pm}0.013$ & $0.540{\pm}0.020$ \\
AudioMNIST-DS         & $0.164{\pm}0.015$ & $0.111{\pm}0.023$ & $0.055{\pm}0.008$ & $0.899{\pm}0.005$ & $0.122{\pm}0.030$ & $0.055{\pm}0.018$ & $0.503{\pm}0.025$ \\
CornellWhaleChallenge & $0.181{\pm}0.010$ & $0.169{\pm}0.003$ & $0.088{\pm}0.001$ & $0.147{\pm}0.054$ & $0.124{\pm}0.003$ & $0.175{\pm}0.004$ & $0.231{\pm}0.002$ \\
CrowdSourced          & $0.338{\pm}0.071$ & $0.337{\pm}0.084$ & $0.328{\pm}0.086$ & $0.338{\pm}0.061$ & $0.349{\pm}0.101$ & $0.315{\pm}0.127$ & $0.341{\pm}0.063$ \\
DREAMERA              & $0.505{\pm}0.045$ & $0.504{\pm}0.030$ & $0.495{\pm}0.023$ & $0.496{\pm}0.013$ & $0.490{\pm}0.039$ & $0.525{\pm}0.017$ & $0.550{\pm}0.050$ \\
DREAMERV              & $0.426{\pm}0.023$ & $0.428{\pm}0.026$ & $0.424{\pm}0.025$ & $0.395{\pm}0.021$ & $0.412{\pm}0.042$ & $0.422{\pm}0.021$ & $0.403{\pm}0.015$ \\
FordChallenge         & $0.100{\pm}0.009$ & $0.098{\pm}0.014$ & $0.107{\pm}0.012$ & $0.104{\pm}0.019$ & $0.069{\pm}0.012$ & $0.216{\pm}0.006$ & $0.076{\pm}0.015$ \\
FruitFlies            & $0.078{\pm}0.010$ & $0.136{\pm}0.003$ & $0.035{\pm}0.003$ & $0.042{\pm}0.014$ & $0.059{\pm}0.002$ & $0.029{\pm}0.001$ & $0.161{\pm}0.004$ \\
InsectSound           & $0.344{\pm}0.008$ & $0.339{\pm}0.014$ & $0.224{\pm}0.007$ & $0.236{\pm}0.009$ & $0.224{\pm}0.004$ & $0.178{\pm}0.006$ & $0.384{\pm}0.008$ \\
LakeIce               & $0.004{\pm}0.000$ & $0.003{\pm}0.001$ & $0.004{\pm}0.000$ & $0.004{\pm}0.000$ & $0.002{\pm}0.000$ & $0.012{\pm}0.001$ & $0.005{\pm}0.000$ \\
LenDB                 & $0.086{\pm}0.021$ & $0.076{\pm}0.016$ & $0.048{\pm}0.011$ & $0.509{\pm}0.013$ & $0.057{\pm}0.014$ & $0.078{\pm}0.022$ & $0.239{\pm}0.040$ \\
MosquitoSound         & $0.156{\pm}0.011$ & $0.188{\pm}0.004$ & $0.076{\pm}0.003$ & $0.082{\pm}0.003$ & $0.173{\pm}0.001$ & $0.131{\pm}0.002$ & $0.442{\pm}0.002$ \\
Opportunity           & $0.184{\pm}0.026$ & $0.167{\pm}0.014$ & $0.177{\pm}0.019$ & $0.181{\pm}0.018$ & $0.134{\pm}0.023$ & $0.181{\pm}0.006$ & $0.147{\pm}0.026$ \\
PAMAP2                & $0.261{\pm}0.117$ & $0.277{\pm}0.116$ & $0.309{\pm}0.110$ & $0.303{\pm}0.100$ & $0.245{\pm}0.118$ & $0.196{\pm}0.095$ & $0.549{\pm}0.067$ \\
Pedestrian            & $0.207{\pm}0.002$ & $0.204{\pm}0.001$ & $0.216{\pm}0.002$ & $0.216{\pm}0.002$ & $0.212{\pm}0.001$ & $0.384{\pm}0.002$ & $0.228{\pm}0.002$ \\
S2Agri-10pc-17        & $0.397{\pm}0.009$ & $0.372{\pm}0.009$ & $0.405{\pm}0.011$ & $0.405{\pm}0.010$ & $0.271{\pm}0.010$ & $0.413{\pm}0.008$ & $0.281{\pm}0.010$ \\
S2Agri-10pc-34        & $0.390{\pm}0.008$ & $0.377{\pm}0.007$ & $0.405{\pm}0.010$ & $0.405{\pm}0.009$ & $0.285{\pm}0.010$ & $0.407{\pm}0.010$ & $0.295{\pm}0.010$ \\
S2Agri-17             & $0.036{\pm}0.001$ & $0.035{\pm}0.001$ & $0.051{\pm}0.002$ & $0.045{\pm}0.001$ & $0.054{\pm}0.001$ & $0.081{\pm}0.001$ & $0.054{\pm}0.001$ \\
S2Agri-34             & $0.043{\pm}0.001$ & $0.041{\pm}0.001$ & $0.058{\pm}0.002$ & $0.053{\pm}0.001$ & $0.061{\pm}0.001$ & $0.090{\pm}0.002$ & $0.061{\pm}0.001$ \\
STEW                  & $0.287{\pm}0.025$ & $0.279{\pm}0.028$ & $0.254{\pm}0.023$ & $0.265{\pm}0.021$ & $0.232{\pm}0.004$ & $0.253{\pm}0.007$ & $0.255{\pm}0.005$ \\
Skoda                 & $0.056{\pm}0.003$ & $0.056{\pm}0.006$ & $0.054{\pm}0.005$ & $0.065{\pm}0.007$ & $0.049{\pm}0.051$ & $0.083{\pm}0.022$ & $0.055{\pm}0.022$ \\
TimeSen2Crop          & $0.043{\pm}0.012$ & $0.047{\pm}0.021$ & $0.040{\pm}0.008$ & $0.042{\pm}0.013$ & $0.210{\pm}0.034$ & $0.409{\pm}0.043$ & $0.216{\pm}0.034$ \\
Tiselac               & $0.288{\pm}0.052$ & $0.284{\pm}0.052$ & $0.286{\pm}0.054$ & $0.289{\pm}0.054$ & $0.219{\pm}0.039$ & $0.251{\pm}0.040$ & $0.223{\pm}0.037$ \\
Traffic               & $0.403{\pm}0.007$ & $0.316{\pm}0.003$ & $0.317{\pm}0.001$ & $0.306{\pm}0.001$ & $0.248{\pm}0.001$ & $0.451{\pm}0.001$ & $0.242{\pm}0.001$ \\
UCIActivity           & $0.094{\pm}0.030$ & $0.067{\pm}0.039$ & $0.061{\pm}0.038$ & $0.077{\pm}0.037$ & $0.080{\pm}0.052$ & $0.038{\pm}0.026$ & $0.093{\pm}0.047$ \\
USCActivity           & $0.325{\pm}0.080$ & $0.314{\pm}0.087$ & $0.343{\pm}0.089$ & $0.340{\pm}0.077$ & $0.278{\pm}0.085$ & $0.333{\pm}0.043$ & $0.315{\pm}0.070$ \\
WISDM                 & $0.187{\pm}0.100$ & $0.186{\pm}0.073$ & $0.155{\pm}0.061$ & $0.152{\pm}0.058$ & $0.117{\pm}0.067$ & $0.151{\pm}0.085$ & $0.189{\pm}0.069$ \\
WISDM2                & $0.271{\pm}0.062$ & $0.267{\pm}0.060$ & $0.291{\pm}0.050$ & $0.301{\pm}0.070$ & $0.263{\pm}0.003$ & $0.271{\pm}0.003$ & $0.276{\pm}0.003$ \\
WhaleSounds           & $0.395{\pm}0.009$ & $0.388{\pm}0.003$ & $0.251{\pm}0.003$ & $0.261{\pm}0.006$ & $0.293{\pm}0.058$ & $0.411{\pm}0.060$ & $0.596{\pm}0.090$ \\
\bottomrule
\end{tabular}

\end{sidewaystable}

\begin{sidewaystable}[htbp]
\centering
\caption{Average error, rank, parameters, and FLOPs per domain on MONSTER}
\label{tab:domain_model}
\tiny
\fontsize{5.5}{6.5}\selectfont
\setlength{\tabcolsep}{2.5pt}
\renewcommand{\arraystretch}{1.1}
\begin{tabular}{llrrrrrrrrrrrrrrr}
\toprule
Domain & Metric & \textbf{MS4N} & \textbf{MS4} & \textbf{ConvTran} & \textbf{Quant} & \textbf{Hydra} & \textbf{Mamba1} & \textbf{Mamba2} & \textbf{GRU} & \textbf{LSTM} & \textbf{HInception} & \textbf{FCN} & \textbf{Informer} & \textbf{Transformer} & \textbf{iTransformer} & \textbf{ET} \\
\midrule
\multirow{4}{*}{\textbf{Audio}} & Error & \textbf{0.1008} & 0.1083 & 0.1214 & 0.1594 & 0.1437 & 0.1796 & 0.1702 & 0.2328 & 0.3668 & 0.1872 & 0.5362 & 0.2151 & 0.2157 & 0.3946 & 0.4081 \\
 & Rank & \textbf{2.29} & 3.86 & 3.71 & 7.57 & 5.43 & 8.57 & 7.86 & 5.43 & 8.57 & 7.00 & 12.29 & 9.86 & 10.43 & 13.29 & 13.57 \\
 & Param & 23,559 & \textbf{23,431} & 29,816 & -- & -- & 55,815 & 39,991 & 52,487 & 69,127 & 74,928 & 269,331 & 285,307 & 285,862 & 731,856 & -- \\
 & FLOPs & 84.99 & 84.37 & 73.66 & -- & -- & 2.49 & 250.47 & 494.64 & 660.06 & 728.49 & 1996.64 & 2805.85 & 2785.87 & \textbf{0.73} & -- \\
\midrule
\multirow{4}{*}{\textbf{EEG}} & Error & 0.3620 & 0.3640 & \textbf{0.3493} & 0.3708 & 0.3788 & 0.3829 & 0.3808 & 0.3754 & 0.3734 & 0.3670 & 0.3790 & 0.3870 & 0.3888 & 0.4231 & 0.3873 \\
 & Rank & \textbf{3.75} & 4.00 & 4.50 & 7.25 & 7.25 & 9.50 & 8.75 & 7.00 & 8.25 & 6.00 & 8.25 & 10.50 & 11.00 & 13.50 & 9.50 \\
 & Param & 24,066 & \textbf{23,938} & 81,786 & -- & -- & 56,322 & 40,498 & 53,154 & 69,794 & 75,552 & 278,259 & 287,478 & 288,255 & 103,874 & -- \\
 & FLOPs & 2.46 & 2.45 & 27.71 & -- & -- & \textbf{0.28} & 6.84 & 13.27 & 17.63 & 19.46 & 53.00 & 73.80 & 72.62 & 1.40 & -- \\
\midrule
\multirow{4}{*}{\textbf{HAR}} & Error & 0.1868 & 0.1885 & 0.1716 & \textbf{0.1667} & 0.1791 & 0.2033 & 0.2056 & 0.1986 & 0.2026 & 0.2020 & 0.1951 & 0.1906 & 0.1970 & 0.3718 & 0.2319 \\
 & Rank & 4.86 & 6.14 & 4.00 & \textbf{3.14} & 6.86 & 10.14 & 9.29 & 8.71 & 11.00 & 8.43 & 7.71 & 7.14 & 9.29 & 14.86 & 9.00 \\
 & Param & 25,828 & \textbf{25,700} & 168,211 & -- & -- & 58,084 & 42,260 & 54,715 & 71,355 & 77,070 & 303,764 & 291,946 & 292,723 & 94,555 & -- \\
 & FLOPs & 1.14 & 1.13 & 26.56 & -- & -- & \textbf{0.25} & 2.91 & 5.53 & 7.30 & 8.04 & 22.91 & 30.29 & 29.91 & 3.15 & -- \\
\midrule
\multirow{4}{*}{\textbf{SI}} & Error & 0.1787 & 0.1782 & \textbf{0.1511} & 0.1574 & 0.2376 & 0.1798 & 0.1789 & 0.1786 & 0.1777 & 0.1842 & 0.1830 & 0.1656 & 0.1715 & 0.2553 & 0.1622 \\
 & Rank & 8.43 & 8.00 & \textbf{2.86} & 5.71 & 13.43 & 8.86 & 9.86 & 8.00 & 8.43 & 9.57 & 8.57 & 3.86 & 5.00 & 11.29 & 7.86 \\
 & Param & 24,740 & \textbf{24,612} & 59,021 & -- & -- & 56,996 & 41,172 & 53,325 & 69,965 & 75,265 & 274,142 & 287,457 & 288,234 & 94,409 & -- \\
 & FLOPs & 0.85 & 0.84 & 6.41 & -- & -- & \textbf{0.07} & 2.42 & 4.74 & 6.31 & 6.96 & 18.15 & 26.41 & 26.02 & 0.76 & -- \\
\midrule
\multirow{4}{*}{\textbf{Other}} & Error & 0.0779 & 0.0830 & 0.3132 & \textbf{0.0626} & 0.1470 & 0.1412 & 0.0862 & 0.0775 & 0.3062 & 0.2757 & 0.3501 & 0.0867 & 0.0929 & 0.1810 & 0.1576 \\
 & Rank & 4.00 & 7.00 & 12.50 & \textbf{2.00} & 10.50 & 9.00 & 8.00 & 5.00 & 10.50 & 8.00 & 14.00 & 4.50 & 7.50 & 11.50 & 6.50 \\
 & Param & 24,226 & \textbf{24,098} & 91,378 & -- & -- & 56,482 & 40,658 & 53,314 & 69,954 & 75,723 & 281,069 & 287,958 & 288,735 & 106,050 & -- \\
 & FLOPs & 2.62 & 2.60 & 11.95 & -- & -- & \textbf{0.15} & 7.57 & 14.87 & 19.80 & 21.87 & 58.16 & 83.30 & 81.75 & 1.47 & -- \\
\midrule
\multirow{4}{*}{\textbf{Counts}} & Error & 0.2806 & 0.2921 & 0.2656 & \textbf{0.2299} & 0.4178 & 0.3008 & 0.3134 & 0.2669 & 0.2609 & 0.2743 & 0.4185 & 0.2600 & 0.3050 & 0.3307 & 0.2348 \\
 & Rank & 6.50 & 10.50 & 3.50 & \textbf{3.00} & 14.50 & 10.50 & 12.50 & 6.50 & 5.00 & 6.50 & 12.00 & \textbf{3.00} & 8.50 & 12.50 & 5.00 \\
 & Param & 25,996 & \textbf{25,868} & 28,276 & -- & -- & 58,252 & 42,428 & 53,724 & 70,364 & 75,848 & 291,829 & 287,744 & 288,133 & 91,788 & -- \\
 & FLOPs & 0.22 & 0.22 & 0.70 & -- & -- & \textbf{0.01} & 0.63 & 1.23 & 1.64 & 1.81 & 4.37 & 6.82 & 6.74 & 0.09 & -- \\
\bottomrule
\end{tabular}
\end{sidewaystable}

\begin{sidewaystable}[htbp]
\centering
\caption{Error rate ${\pm}$ STD per dataset and model on UEA benchmark}
\label{tab:error_rate_uea}
\tiny
\fontsize{6}{7}\selectfont
\setlength{\tabcolsep}{3pt}

\begin{tabular}{lrrrrrr}
\toprule
Dataset & MS4N & MS4 & Mamba1 & Mamba2 & ConvTran & HInception \\
\midrule
ArticularyWordRecognition & $0.035{\pm}0.009$ & $0.034{\pm}0.015$ & $0.138{\pm}0.022$ & $0.051{\pm}0.017$ & $0.016{\pm}0.006$ & $0.024{\pm}0.008$ \\
AtrialFibrillation        & $0.677{\pm}0.081$ & $0.687{\pm}0.070$ & $0.818{\pm}0.084$ & $0.742{\pm}0.090$ & $0.787{\pm}0.074$ & $0.707{\pm}0.067$ \\
BasicMotions              & $0.000{\pm}0.000$ & $0.002{\pm}0.012$ & $0.029{\pm}0.035$ & $0.000{\pm}0.000$ & $0.002{\pm}0.007$ & $0.001{\pm}0.004$ \\
CharacterTrajectories     & $0.006{\pm}0.002$ & $0.008{\pm}0.002$ & $0.018{\pm}0.005$ & $0.019{\pm}0.006$ & $0.006{\pm}0.002$ & $0.006{\pm}0.004$ \\
Cricket                   & $0.019{\pm}0.008$ & $0.019{\pm}0.009$ & $0.062{\pm}0.027$ & $0.030{\pm}0.014$ & $0.015{\pm}0.013$ & $0.014{\pm}0.002$ \\
DuckDuckGeese             & $0.408{\pm}0.040$ & $0.412{\pm}0.072$ & $0.452{\pm}0.062$ & $0.491{\pm}0.048$ & $0.492{\pm}0.081$ & $0.445{\pm}0.047$ \\
ERing                     & $0.144{\pm}0.031$ & $0.179{\pm}0.131$ & $0.278{\pm}0.078$ & $0.177{\pm}0.040$ & $0.105{\pm}0.124$ & $0.160{\pm}0.077$ \\
EigenWorms                & $0.139{\pm}0.041$ & $0.179{\pm}0.052$ & $0.193{\pm}0.035$ & $0.166{\pm}0.038$ & $0.346{\pm}0.075$ & $0.594{\pm}0.057$ \\
Epilepsy                  & $0.061{\pm}0.012$ & $0.062{\pm}0.015$ & $0.071{\pm}0.023$ & $0.033{\pm}0.011$ & $0.026{\pm}0.012$ & $0.028{\pm}0.008$ \\
EthanolConcentration      & $0.704{\pm}0.023$ & $0.709{\pm}0.021$ & $0.707{\pm}0.024$ & $0.711{\pm}0.023$ & $0.698{\pm}0.019$ & $0.708{\pm}0.016$ \\
FaceDetection             & $0.375{\pm}0.009$ & $0.365{\pm}0.010$ & $0.454{\pm}0.012$ & $0.464{\pm}0.010$ & $0.343{\pm}0.009$ & $0.358{\pm}0.013$ \\
FingerMovements           & $0.504{\pm}0.036$ & $0.506{\pm}0.027$ & $0.504{\pm}0.028$ & $0.488{\pm}0.046$ & $0.443{\pm}0.047$ & $0.448{\pm}0.043$ \\
HandMovementDirection     & $0.623{\pm}0.051$ & $0.613{\pm}0.071$ & $0.709{\pm}0.045$ & $0.683{\pm}0.062$ & $0.621{\pm}0.060$ & $0.627{\pm}0.050$ \\
Handwriting               & $0.643{\pm}0.073$ & $0.747{\pm}0.132$ & $0.732{\pm}0.042$ & $0.717{\pm}0.049$ & $0.603{\pm}0.052$ & $0.520{\pm}0.041$ \\
Heartbeat                 & $0.274{\pm}0.029$ & $0.254{\pm}0.018$ & $0.283{\pm}0.032$ & $0.276{\pm}0.026$ & $0.268{\pm}0.039$ & $0.285{\pm}0.043$ \\
InsectWingbeat            & $0.303{\pm}0.004$ & $0.311{\pm}0.006$ & $0.381{\pm}0.005$ & $0.382{\pm}0.006$ & $0.286{\pm}0.005$ & $0.313{\pm}0.004$ \\
JapaneseVowels            & $0.023{\pm}0.006$ & $0.025{\pm}0.004$ & $0.047{\pm}0.012$ & $0.022{\pm}0.005$ & $0.018{\pm}0.005$ & $0.015{\pm}0.006$ \\
LSST                      & $0.340{\pm}0.016$ & $0.363{\pm}0.026$ & $0.401{\pm}0.024$ & $0.339{\pm}0.009$ & $0.371{\pm}0.018$ & $0.465{\pm}0.023$ \\
Libras                    & $0.155{\pm}0.023$ & $0.154{\pm}0.027$ & $0.291{\pm}0.038$ & $0.252{\pm}0.037$ & $0.108{\pm}0.022$ & $0.132{\pm}0.024$ \\
MotorImagery              & $0.480{\pm}0.048$ & $0.477{\pm}0.044$ & $0.497{\pm}0.035$ & $0.496{\pm}0.033$ & $0.480{\pm}0.042$ & $0.482{\pm}0.042$ \\
NATOPS                    & $0.066{\pm}0.025$ & $0.089{\pm}0.055$ & $0.068{\pm}0.023$ & $0.059{\pm}0.019$ & $0.054{\pm}0.017$ & $0.050{\pm}0.016$ \\
PEMS-SF                   & $0.181{\pm}0.032$ & $0.179{\pm}0.026$ & $0.189{\pm}0.033$ & $0.227{\pm}0.032$ & $0.224{\pm}0.029$ & $0.261{\pm}0.034$ \\
PenDigits                 & $0.016{\pm}0.003$ & $0.016{\pm}0.002$ & $0.018{\pm}0.004$ & $0.021{\pm}0.003$ & $0.014{\pm}0.004$ & $0.016{\pm}0.003$ \\
PhonemeSpectra            & $0.731{\pm}0.014$ & $0.740{\pm}0.015$ & $0.772{\pm}0.008$ & $0.746{\pm}0.009$ & $0.708{\pm}0.007$ & $0.731{\pm}0.014$ \\
RacketSports              & $0.143{\pm}0.021$ & $0.164{\pm}0.028$ & $0.239{\pm}0.028$ & $0.164{\pm}0.027$ & $0.160{\pm}0.022$ & $0.135{\pm}0.021$ \\
SelfRegulationSCP1        & $0.164{\pm}0.037$ & $0.168{\pm}0.058$ & $0.241{\pm}0.068$ & $0.215{\pm}0.033$ & $0.139{\pm}0.032$ & $0.192{\pm}0.083$ \\
SelfRegulationSCP2        & $0.481{\pm}0.027$ & $0.478{\pm}0.029$ & $0.480{\pm}0.031$ & $0.485{\pm}0.028$ & $0.474{\pm}0.036$ & $0.470{\pm}0.026$ \\
SpokenArabicDigits        & $0.006{\pm}0.002$ & $0.008{\pm}0.004$ & $0.027{\pm}0.006$ & $0.035{\pm}0.007$ & $0.008{\pm}0.005$ & $0.008{\pm}0.004$ \\
StandWalkJump             & $0.680{\pm}0.063$ & $0.682{\pm}0.070$ & $0.670{\pm}0.085$ & $0.683{\pm}0.109$ & $0.692{\pm}0.095$ & $0.665{\pm}0.081$ \\
UWaveGestureLibrary       & $0.101{\pm}0.025$ & $0.091{\pm}0.021$ & $0.229{\pm}0.056$ & $0.200{\pm}0.058$ & $0.100{\pm}0.018$ & $0.111{\pm}0.016$ \\
\bottomrule
\end{tabular}

\vspace{4pt}

\begin{tabular}{lrrrrrr}
\toprule
Dataset & FCN & Informer & Transformer & iTransformer & GRU & LSTM \\
\midrule
ArticularyWordRecognition & $0.062{\pm}0.013$ & $0.024{\pm}0.009$ & $0.038{\pm}0.013$ & $0.218{\pm}0.028$ & $0.100{\pm}0.019$ & $0.174{\pm}0.029$ \\
AtrialFibrillation        & $0.783{\pm}0.065$ & $0.685{\pm}0.080$ & $0.772{\pm}0.085$ & $0.718{\pm}0.086$ & $0.672{\pm}0.071$ & $0.685{\pm}0.069$ \\
BasicMotions              & $0.000{\pm}0.000$ & $0.000{\pm}0.000$ & $0.020{\pm}0.025$ & $0.444{\pm}0.056$ & $0.069{\pm}0.046$ & $0.078{\pm}0.043$ \\
CharacterTrajectories     & $0.028{\pm}0.006$ & $0.014{\pm}0.004$ & $0.014{\pm}0.004$ & $0.026{\pm}0.004$ & $0.017{\pm}0.005$ & $0.094{\pm}0.144$ \\
Cricket                   & $0.016{\pm}0.005$ & $0.028{\pm}0.013$ & $0.026{\pm}0.021$ & $0.319{\pm}0.030$ & $0.066{\pm}0.017$ & $0.102{\pm}0.034$ \\
DuckDuckGeese             & $0.402{\pm}0.053$ & $0.418{\pm}0.063$ & $0.442{\pm}0.077$ & $0.810{\pm}0.027$ & $0.486{\pm}0.066$ & $0.516{\pm}0.052$ \\
ERing                     & $0.200{\pm}0.040$ & $0.114{\pm}0.023$ & $0.160{\pm}0.048$ & $0.206{\pm}0.078$ & $0.235{\pm}0.026$ & $0.254{\pm}0.039$ \\
EigenWorms                & $0.191{\pm}0.035$ & $0.206{\pm}0.033$ & $0.302{\pm}0.048$ & $0.616{\pm}0.040$ & $0.543{\pm}0.048$ & $0.556{\pm}0.029$ \\
Epilepsy                  & $0.045{\pm}0.015$ & $0.029{\pm}0.008$ & $0.046{\pm}0.013$ & $0.572{\pm}0.063$ & $0.066{\pm}0.024$ & $0.139{\pm}0.043$ \\
EthanolConcentration      & $0.717{\pm}0.018$ & $0.724{\pm}0.018$ & $0.727{\pm}0.022$ & $0.747{\pm}0.014$ & $0.702{\pm}0.028$ & $0.702{\pm}0.027$ \\
FaceDetection             & $0.457{\pm}0.009$ & $0.354{\pm}0.009$ & $0.381{\pm}0.015$ & $0.444{\pm}0.007$ & $0.402{\pm}0.010$ & $0.413{\pm}0.013$ \\
FingerMovements           & $0.505{\pm}0.040$ & $0.490{\pm}0.035$ & $0.486{\pm}0.037$ & $0.456{\pm}0.039$ & $0.476{\pm}0.046$ & $0.473{\pm}0.033$ \\
HandMovementDirection     & $0.642{\pm}0.039$ & $0.429{\pm}0.059$ & $0.645{\pm}0.069$ & $0.606{\pm}0.040$ & $0.656{\pm}0.040$ & $0.636{\pm}0.045$ \\
Handwriting               & $0.664{\pm}0.063$ & $0.729{\pm}0.050$ & $0.752{\pm}0.035$ & $0.860{\pm}0.024$ & $0.876{\pm}0.062$ & $0.915{\pm}0.017$ \\
Heartbeat                 & $0.265{\pm}0.030$ & $0.283{\pm}0.042$ & $0.276{\pm}0.027$ & $0.337{\pm}0.054$ & $0.333{\pm}0.028$ & $0.344{\pm}0.037$ \\
InsectWingbeat            & $0.369{\pm}0.005$ & $0.379{\pm}0.004$ & $0.411{\pm}0.004$ & $0.731{\pm}0.003$ & $0.405{\pm}0.006$ & $0.489{\pm}0.174$ \\
JapaneseVowels            & $0.023{\pm}0.004$ & $0.020{\pm}0.009$ & $0.019{\pm}0.012$ & $0.344{\pm}0.030$ & $0.033{\pm}0.012$ & $0.046{\pm}0.011$ \\
LSST                      & $0.338{\pm}0.016$ & $0.379{\pm}0.015$ & $0.386{\pm}0.014$ & $0.515{\pm}0.013$ & $0.363{\pm}0.012$ & $0.380{\pm}0.014$ \\
Libras                    & $0.252{\pm}0.033$ & $0.171{\pm}0.030$ & $0.171{\pm}0.030$ & $0.453{\pm}0.032$ & $0.181{\pm}0.028$ & $0.299{\pm}0.037$ \\
MotorImagery              & $0.504{\pm}0.032$ & $0.490{\pm}0.037$ & $0.481{\pm}0.045$ & $0.501{\pm}0.046$ & $0.520{\pm}0.034$ & $0.508{\pm}0.043$ \\
NATOPS                    & $0.048{\pm}0.014$ & $0.088{\pm}0.023$ & $0.066{\pm}0.020$ & $0.380{\pm}0.045$ & $0.066{\pm}0.030$ & $0.086{\pm}0.043$ \\
PEMS-SF                   & $0.235{\pm}0.032$ & $0.178{\pm}0.030$ & $0.238{\pm}0.035$ & $0.211{\pm}0.039$ & $0.243{\pm}0.038$ & $0.281{\pm}0.249$ \\
PenDigits                 & $0.021{\pm}0.003$ & $0.019{\pm}0.004$ & $0.014{\pm}0.003$ & $0.024{\pm}0.003$ & $0.014{\pm}0.003$ & $0.016{\pm}0.004$ \\
PhonemeSpectra            & $0.773{\pm}0.010$ & $0.775{\pm}0.007$ & $0.780{\pm}0.008$ & $0.918{\pm}0.004$ & $0.786{\pm}0.009$ & $0.817{\pm}0.008$ \\
RacketSports              & $0.183{\pm}0.018$ & $0.161{\pm}0.018$ & $0.196{\pm}0.037$ & $0.362{\pm}0.044$ & $0.179{\pm}0.039$ & $0.189{\pm}0.036$ \\
SelfRegulationSCP1        & $0.227{\pm}0.067$ & $0.174{\pm}0.041$ & $0.164{\pm}0.027$ & $0.190{\pm}0.042$ & $0.364{\pm}0.060$ & $0.393{\pm}0.068$ \\
SelfRegulationSCP2        & $0.472{\pm}0.022$ & $0.485{\pm}0.022$ & $0.478{\pm}0.028$ & $0.498{\pm}0.024$ & $0.482{\pm}0.031$ & $0.499{\pm}0.032$ \\
SpokenArabicDigits        & $0.048{\pm}0.004$ & $0.016{\pm}0.005$ & $0.014{\pm}0.005$ & $0.074{\pm}0.008$ & $0.049{\pm}0.138$ & $0.899{\pm}0.008$ \\
StandWalkJump             & $0.668{\pm}0.082$ & $0.740{\pm}0.096$ & $0.670{\pm}0.084$ & $0.572{\pm}0.080$ & $0.663{\pm}0.083$ & $0.665{\pm}0.090$ \\
UWaveGestureLibrary       & $0.414{\pm}0.034$ & $0.135{\pm}0.017$ & $0.199{\pm}0.036$ & $0.340{\pm}0.017$ & $0.281{\pm}0.034$ & $0.474{\pm}0.037$ \\
\bottomrule
\end{tabular}

\end{sidewaystable}

\end{document}